\pdfoutput=1
\documentclass[11pt]{article}

\usepackage[final]{acl}

\usepackage{times}
\usepackage{latexsym}

\usepackage[T1]{fontenc}
\usepackage[utf8]{inputenc}

\usepackage{microtype}

\usepackage{inconsolata}

\usepackage{graphicx}

\usepackage{amsmath}
\usepackage{booktabs}      % \toprule \midrule \bottomrule
\usepackage{tabularx}      
\usepackage{makecell}      
\usepackage{pifont}        
\usepackage{hyperref}  
\usepackage{subcaption}

\newcommand{\cmark}{\ding{51}} % ✓
\newcommand{\xmark}{\ding{55}} % ✗
\usepackage{tabularx}  
\usepackage{multirow}
\usepackage{tcolorbox}
\tcbuselibrary{skins, breakable}

\title{ObjChangeVR: Object State Change Reasoning from Continuous Egocentric Views in VR Environments}

\author{
  Shiyi Ding\textsuperscript{1},
  Shaoen Wu\textsuperscript{1},
  Ying Chen\textsuperscript{2} \\
  \textsuperscript{1}Kennesaw State University,
  \textsuperscript{2}Pennsylvania State University \\
  \texttt{sding1@students.kennesaw.edu, swu10@kennesaw.edu, yingchen@psu.edu}
}

\newcommand{\name}{ObjChangeVR}

\begin{document}
\maketitle
\begin{abstract}
Recent advances in multimodal large language models (MLLMs) offer a promising approach for natural language-based scene change queries in virtual reality (VR).
Prior work on applying MLLMs for object state understanding has focused on egocentric videos that capture the camera wearer's interactions with objects. 
However, object state changes may occur in the background without direct user interaction, lacking explicit motion cues and making them difficult to detect. Moreover, no benchmark exists for evaluating this challenging scenario.
 To address these challenges, we introduce ObjChangeVR-Dataset, specifically for benchmarking the question-answering task of object state change. We also propose ObjChangeVR, a framework that combines viewpoint-aware and temporal-based retrieval to identify relevant frames, along with cross-view reasoning that reconciles inconsistent evidence from multiple viewpoints. Extensive experiments demonstrate that ObjChangeVR significantly outperforms baseline approaches across multiple MLLMs.

\end{abstract}

\section{Introduction}

Virtual reality (VR) has attracted increasing attention in various fields, such as entertainment, social interactions, and commerce~\cite{fortunebusinessinsights_vr_market}. 
As VR environments become more dynamic and interactive, accurately identifying and localizing scene changes between past and present user views has become an essential task for 3D scene understanding, supporting diverse applications ranging from interactive training simulations~\cite{bjorn2024social} to collaborative virtual workspaces~\cite{jing2023visualising}.

\begin{figure}[t]
    \centering
\includegraphics[width=0.49\textwidth]{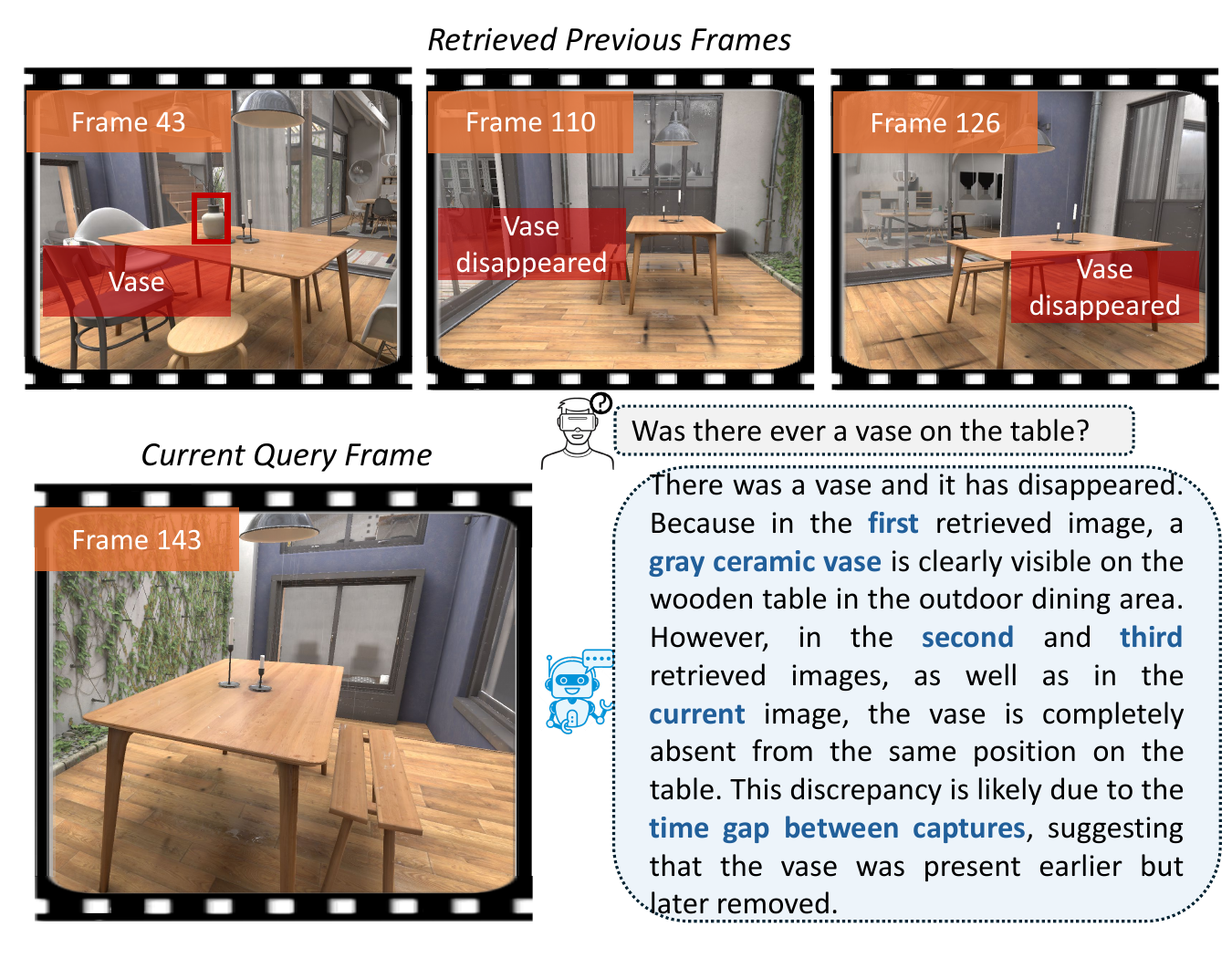}
    \caption{Illustration of the question-answering task for object state change reasoning. Given a query frame and a 
    question about object change, we retrieve several relevant frames from the egocentric frame sequence and leverage visual evidence from the retrieved frames to produce an answer and an explanation.}
    \label{fig:task}
    \vspace{-0.3cm}
\end{figure}

Recent advances in multimodal large language models (MLLMs) have shown remarkable capabilities in 
scene understanding~\citep{gpt4v, gemini, qwen-vl, liu2023visual},  offering a promising approach for natural language-based scene change queries in VR environments.
Although promising, 
current scene change detection methods still 
mainly rely on computer vision techniques that localize pixel-wise change regions~\citep{sachdeva2023change,lee2024semi,furukawa2020self}, lacking support for natural language queries, which allow for a more intuitive and flexible interaction modality for VR users.

Directly applying MLLMs to VR scene change detection faces three challenges. First, 
users traverse VR environments, generating lengthy frame sequences, but only a small subset contains evidence relevant to a given query on scene change. It is challenging to identify which frames are informative while processing long input sequences. Second, unlike existing egocentric video-based benchmarks~\cite{xiao2021next,di2024grounded,ye2024mm}, 
object state changes might occur in the background without direct human interaction. For instance, objects may be moved or reconfigured (by another VR user) while a user explores distant areas. Since these background changes lack explicit motion cues and exhibit low perceptual saliency, they are 
more challenging to detect than changes caused by direct user manipulation captured in egocentric views. Third, no existing benchmark dataset evaluates natural language-based object state change reasoning in continuous egocentric views where user viewpoints shift dramatically across scene sections (e.g., from a kitchen to a study room).

To address these challenges, we introduce a question-answering task on object state change reasoning in continuous egocentric video streams, as shown in Figure~\ref{fig:task}. We consider a user navigating through an environment, potentially moving across different scene sections 
and returning to previously visited sections. During this navigation, egocentric frames are continuously captured, and object states may change over time (%not a result of direct user interaction and 
the process of object state change is not captured in egocentric videos). Given a natural language question such as \textit{``Was there ever a vase on the table?''}, the task involves reasoning over multiple previous frames to determine whether an object ever existed.

We collect a dataset, called \name{}-Dataset, specifically for benchmarking the object state change question-answering task, and propose a framework, \name{}, to tackle this task. 

First, to 
identify informative frames from lengthy continuous egocentric frames,  \name{} leverages viewpoint metadata provided by VR devices to retrieve frames 
containing relevant visual evidence for answering object state change queries.
Such viewpoint metadata is increasingly accessible in cameras. 
Extended reality headsets (e.g., Meta Quest series~\cite{metaquest}, Apple Vision Pro~\cite{applevisionpro}) automatically record  6-degree-of-freedom (6-DoF) camera pose, and many other consumer devices, such as smartphones (both Android and iOS) or cameras (e.g., GoPro~\cite{gopro}, RealSense~\cite{realsenseai}), can capture visual-inertial and depth sensor data, which can be used to compute camera pose through standard simultaneous localization and mapping (SLAM) pipelines~\cite{cadena2017past}, where the pose metadata is then accessible through developer SDK. Therefore, although our current experiments focus on VR, the proposed framework can be applied to real-world egocentric videos when pose information is available or can be reconstructed. 

Second, to combat the challenge that object changes occurring without direct user interaction lack explicit motion cues and exhibit low perceptual saliency, \name{} adopts cross-view reasoning over retrieved frames to improve detection accuracy. 
Since the retrieved 
frames are captured from different viewpoints and at different times, they vary in informativeness about the object’s state. \name{} prioritizes %those with better 
viewpoints with higher informativeness for more reliable reasoning.  
\name{} also 
guides reasoning across temporally ordered frames. For example, when an object consistently appears in earlier frames but is absent in later frames, this pattern provides strong evidence for disappearance rather than mere occlusion. By treating cross-frame inconsistencies as cues rather than noise, our approach effectively distinguishes genuine object state changes from transient observation artifacts.

The main contributions of our paper are: 1)
We introduce \name{}-Dataset, a benchmark for object state change reasoning in continuous egocentric views. The dataset comprises 5 diverse VR scenes (e.g., villa interior, outdoor market) spanning 35 distinct scene sections (e.g., first-floor kitchen, fish shop), with 729 target objects whose states may change over time. 2) We propose \name{}, which combines viewpoint-aware 
relevant frame retrieval 
with a cross-view reasoning module that aggregates and reconciles inconsistent answers from multiple viewpoints to achieve more accurate object change detection. 
3) Through 
experiments, we demonstrate that \name{} outperforms baseline approaches for frame retrieval and cross-frame inconsistency resolution 
across multiple MLLMs. Our dataset and code are publicly available at:~\url{https://github.com/sding11/ObjChangeVR}.

\section{Related Work}

\noindent\textbf{Scene change detection.} 
Scene change detection aims to identify and localize differences between two observations of the same scene captured at different times. Prior studies have leveraged convolutional neural networks (CNNs) to detect changed regions~\cite{noman2024elgc,han2023hanet,yu2024maskcd,dong2025remote}. However, they assume nearly perfect image alignment between the compared frames. 
While some recent works have addressed unaligned image pairs~\cite{sachdeva2023change,lee2024semi,furukawa2020self}, they still focus on static image pairs captured from similar viewpoints and do not leverage viewpoint metadata from the camera. In contrast, our task involves analyzing sequences of egocentric frames that exhibit frequent and substantial viewpoint shifts. We incorporate explicit viewpoint metadata available from VR devices to reason about spatial relationships across frames. Moreover, rather than delineating change areas through traditional computer vision techniques, we frame the problem as visual question answering to enable more natural interaction for VR users, offering a new angle on scene change detection that remains largely unexplored.

\noindent \textbf{3D and video-based question answering.} 
Prior work~\cite{wu20243d,yan2023comprehensive,ma2022sqa3d,yan2023comprehensive} on 3D question answering has 
leveraged 3D scans 
from datasets like ScanNet~\cite{Scannet} and focused on
processing point cloud data from 
3D 
scenes to respond to specific textual queries about the scenes. A few works have explored using MLLMs for 3D VR scene understanding via situated 3D question answering~\cite{wan2024building,wu2023simmc,ding2025rag,li2025exploring}. 
However, these methods mainly focus on interactive 3D scenes where object changes (if any) are caused by users' own direct interactions, rather than object changes that occur in the background. 
Apart from 3D question answering, another line of work addresses video-based question answering~\cite{mogrovejo2024question,pan2023retrieving,song2025moviechat+}. 
A notable subset of this research focuses on natural language queries in egocentric videos~\cite{xiao2021next,di2024grounded,ye2024mm}, examining human activities to evaluate models' ability to interpret complex actions and interactions.
Complementary to these works, we address object state changes that occur in the background without direct human interaction and without being captured in video recordings. Since these changes lack explicit motion cues and exhibit low perceptual saliency, they present a more challenging question-answering task.

\noindent \textbf{3D and video-based benchmarks for scene understanding.} 
3D scene datasets such as ScanQA~\citep{ScanQA}, MMScan~\citep{MMScan}, VLA-3D~\citep{VLA-3D},  and SIMMC-VR~\citep{wu2023simmc} mainly feature static 3D scans without temporal dynamics.  
Perhaps closest to our work is ChangeSim~\citep{ChangeSim}, which is a benchmark for object state change in a virtual environment. 
While sequences are captured, ChangeSim operates on paired frames at two discrete timestamps and focuses on annotating pixel-level change detection. In contrast, our collected dataset focuses on natural language queries over extended trajectories.
Video-based QA benchmarks such as Causal-VidQA~\citep{Causal-VidQA}, FunQA~\citep{FunQA}, SurveillanceVQA-589K~\citep{SurveillanceVQA-589K}, Pano-AVQA~\citep{Pano-AVQA}, and VideoEspresso~\citep{VideoEspresso} focus on real-world clips that evaluate event understanding, 
where the majority of events last for $5-20$s. 
Egocentric datasets such as Ego4D~\citep{Ego4D}, EPIC-KITCHENS~\citep{damen2021rescaling}, and EgoTracks~\citep{tang2023egotracks} capture human actions, interactions, and object manipulations from a first-person perspective. However, these datasets typically focus on confined scene sections (e.g., a kitchen), where object changes primarily result from direct human interaction.
In contrast, \name{}-Dataset contains 
trajectories traversing multiple scene sections with drastic viewpoint changes. Additionally, object state changes occur without direct human interaction, meaning that objects may be altered outside the user's immediate view or control.
We summarize the differences of our dataset and existing datasets in Table~\ref{tab:dataset_comparison} in Appendix~\ref{sec:dataset-comparision}.

\section{\name{}-Dataset}

\begin{figure*}[t]
    \centering
    \includegraphics[width=0.99\textwidth]{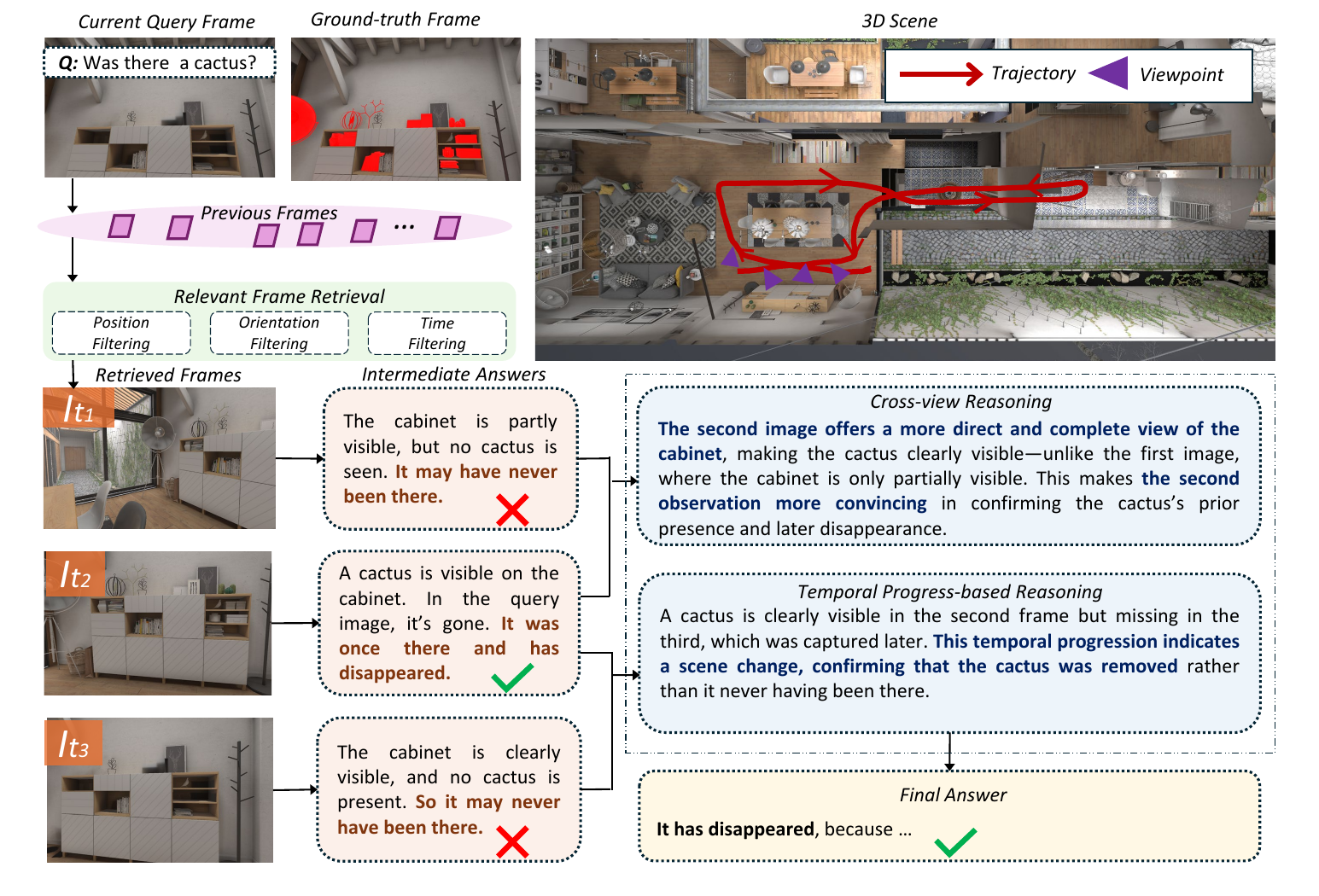}
       \vspace{-0.3cm}
    \caption{Overview of the \name{}-Dataset and the proposed \name{} framework.}
  
    \label{fig:method}
    \vspace{-0.3cm}
\end{figure*}

\begin{table}[t]
\setlength{\tabcolsep}{3.5pt}
\centering
\small
%\vspace{-0.3cm}
\begin{tabular}{cccc}
\hline
{VR scene} & {Scale} &  \# scene sections& \# target objects\\
\hline
Villa interior     & Small & 9 & 421 \\
Restaurant         & Small & 7 & 62  \\
Market             & Small & 5 & 52  \\
Museum             & Large & 6 & 134 \\
Viking village     & Large & 8 & 60  \\
\hline
\end{tabular}
\caption{Statistics of %VR scenes in the 
\name{}-Dataset.}
 \vspace{-0.5cm}
\label{tab:data-stats}
\end{table}
\noindent\textbf{Scene statistics.} \name{}-Dataset uses five distinct VR scenes from the Unity Asset Store, including 4 indoor (\texttt{Villa interior}~\cite{ArchVizPRO}, \texttt{Restaurant}~\cite{FastFood}, \texttt{Market}~\cite{Market}, \texttt{Museum}~\cite{museum}), and 1 outdoor (\texttt{Viking
village}~\cite{VikingVillage15}) scenes. 
We navigate through each VR scene, mimicking natural exploration in VR.
The trajectory is recorded at a frequency of $5$Hz, where the camera’s position and orientation are captured continuously over time. Egocentric images are saved at $1$Hz. As shown in Table~\ref{tab:data-stats}, there are 35 scene sections in total (e.g., first-floor kitchen, fish shop, blacksmith workshop). 
Please see Appendix~\ref{sec:datasetdetail} for details.

\noindent\textbf{Question statistics.}
Our dataset is divided into two categories based on trajectory length.
The \emph{short-trajectory} data comprises approximately $60$s walks in three smaller scenes, \texttt{villa interior}, \texttt{restaurant}, and \texttt{market}, with a total of 3,000 questions related to object state change. The \emph{long-trajectory} data comprises approximately $180$s walks in two larger scenes, \texttt{museum} and \texttt{Viking village}, with a total of 2,000 questions. Long trajectories cover larger scene scales and temporal variations than short trajectories. 

\noindent \textbf{Statistics of target objects.} Our question-answering task requires determining whether target objects have changed their states by the time their scene section is revisited and they are observed from different perspectives. 
The target objects vary in scale and complexity, including large objects (e.g., dinosaur skeletons, barrel stacks), small items (e.g., vases, photo frames, cookies), and grouped objects that disappear together (e.g., tables with their associated items). 
Across all scenes, there are 729 target objects in total, as shown in Table~\ref{tab:data-stats}.

\noindent\textbf{User trajectories.} Data is recorded automatically as we traverse the virtual environments (Figure.~\ref{fig:method} shows an example of a 3D scene and a trajectory), capturing target objects scattered across the scene through an egocentric view. As the trajectory continues, previously seen objects naturally leave the camera’s field of view as we explore new scene sections.
At a specific point during the walk, certain objects change their state (e.g., disappear) while others remain unchanged. 
Users then revisit the same scene sections, but without retracing their exact steps. Instead, they return along different routes with varied angles and distances to the target objects. This means that the objects are observed from new perspectives, for example, passing an area from the opposite direction or at a different distance.  Although these viewpoints still include the areas where the objects were once observed, the positional and angular differences of viewpoints make it difficult to align observations across time. This setup creates a realistic temporal reasoning challenge, allowing us to evaluate whether models can determine object state changes over time while handling viewpoint variation between current frame and previous frames.

\noindent\textbf{Annotations.} We develop a semi-automated pipeline for ground-truth annotations,  which leverages both the Unity game engine (which generates object-level masks) and the reasoning capabilities of MLLMs, followed by human verification. 
We use Unity to overlay object masks directly onto ground-truth frames (the ground-truth frame is shown in Figure~\ref{fig:method}), highlighting objects that have changed their state in a high-contrast color (e.g., red). These annotated images, along with explicit instructions that the red-highlighted regions correspond to objects that were previously present but have now disappeared, are then provided to an MLLM (GPT-4o)
to generate ground-truth answers to questions about 
objects' state change.
Finally, human reviewers verify and correct the generated annotations to ensure correctness.

\section{Methodology}

Consider a user navigating through an environment, potentially moving across different scene sections and returning to previously visited sections. During this navigation, egocentric frames are continuously captured, and object states may change over time. 
 Let $H_t = \{I_i \mid i < t\}$ denote the set of all frames captured before the current timestamp $t$, where $I_i$ is the frame recorded at time $i$. Let $I_c$ denote the current egocentric frame. Each frame is associated with the camera's viewpoint location ($x$, $y$, and $z$ coordinates), orientation (represented as a quaternion ($q_w$, $q_x$, $q_y$, $q_z$)), and timestamp.
Given a current frame $I_c$ 
and a natural language question $\mathcal{Q}$ about the state of a specific object (e.g., whether the object has disappeared, has always been present, or was never there), 
the goal is to generate a natural language answer and explanation indicating the object’s state change. To support this temporal reasoning task, we retrieve a set of previous egocentric frames from the past frame set $H_t$ (Section~\ref{sec:Relevant Cross-view Frame Retrieval}). 
Specifically, we select up to $k$ past frames $\{I_{t_j}\}_{j=1}^{k} \subseteq H_t$ that are most relevant to the question $\mathcal{Q}$, i.e., most likely to contain information that helps determine the object’s state, where timestamps $t_1<t_2< \ldots<t_k<t$ indicate the chronological order in which these past frames were captured. 
These selected frames, along with $I_c$ and $\mathcal{Q}$, are provided to the MLLM as additional context to generate the final answer (Section~\ref{sec:Temporal Cross-View CoT}).
$k$ is a configurable hyperparameter, which is chosen based on computational cost or limits on MLLM input size (e.g., token budget).

\subsection{Relevant Cross-view Frame Retrieval}
\label{sec:Relevant Cross-view Frame Retrieval}
A natural approach to retrieving relevant frames 
is to compare the visual embeddings of past frames with the current query frame $I_c$.
However, this  
may return misleading matches that look similar but come from different areas.  
For instance, hallways 
may have the same wall colors, flooring, and chairs arranged in similar ways. 
To address this, we leverage sensor data automatically recorded by VR devices, including the camera position and orientation associated with each frame. 
This ensures that retrieved frames not only share visual similarity but also originate from the correct spatial region.

To retrieve relevant past frames, we apply a three-stage hierarchical filtering: 
position filtering,   orientation filtering, and
temporal filtering.
First, \emph{position filtering} selects the top $k_p$ previous frames whose viewpoint positions have the smallest Euclidean distance to the position $\mathbf{p}_c$ of the current frame $I_C$, ensuring spatial proximity. Among these frames, \emph{orientation filtering} further keeps the top $k_o$ frames whose viewpoint orientations are most closely aligned with the orientation (represented as a quaternion $\mathbf{q}_c$) of the current frame $I_c$, favoring similar viewing angles.
Finally, \emph{temporal filtering} selects the earliest $k$ frames from the orientation-filtered set to maintain chronological diversity.

Next, we dynamically adjust cutoff values $k_p$ and $k_o$ for the position-filtered and orientation-filtered sets based on how many relevant frames the system ultimately needs to select.
If $k$ increases but $k_p$ and $k_o$ remain small, relevant frames may be prematurely discarded, resulting in low retrieval recall. Conversely, if $k_p$ and $k_o$ are too large, the intermediate filters may admit too many redundant candidates, diminishing the effectiveness of the final filter. To balance retrieval precision and recall, we grow the position-filtered and orientation-filtered sets proportionally to $k$, while bounding their sizes within configurable minimum and maximum limits to avoid overly narrow or 
large selections:
{
\begin{equation*}
\begin{aligned}
k_o &= \min\!\Big(|H_t|,\;\mathrm{cap}_o,\;\max\big(\mathrm{min}_o,\;\lceil \alpha\,k\rceil\big)\Big),\\
k_p &= \min\!\Big(|H_t|,\;\mathrm{cap}_p,\;\max\big(\mathrm{min}_p,\;\lceil \beta\,k_o\rceil\big)\Big),
\end{aligned}
\end{equation*}
}where $\alpha$ and $\beta$ are linear scaling factors that control the size of the orientation- and position-filtered sets relative to the final budget $k$, 
$\mathrm{min}_o$ and $\mathrm{min}_p$ ensure the filters are not overly restrictive when $k$ is small, and $\mathrm{cap}_o$ and $\mathrm{cap}_p$ limit the set sizes to avoid excessive growth when $k$ is large.

\subsection{Temporal Cross-view Reasoning}
\label{sec:Temporal Cross-View CoT}

After retrieving $k$ previous frames $\{I_{t_j}\}_{j=1}^{k}$ that are likely to provide relevant context for answering an object state change question, we aim to let the MLLM output the final answer about the object state change. Our strategy incorporates temporal cross-view information and  exploits a two-stage chain-of-thought (CoT) prompting that 1) 
performs pairwise reasoning between each retrieved frame and the current frame to generate $k$ intermediate answers, and 2) derives the final answer by aggregating and reconciling these intermediate answers when inconsistencies arise. 

\noindent\textbf{Independent intermediate answers.} We provide few-shot exemplars that guide the MLLM
through comparing each retrieved frame and the current frame,
describing salient visual differences, and answering whether the object has changed. This process yields independent $k$ intermediate answers, each accompanied by a brief explanation that combines information from both the current frame and the corresponding retrieved frame to justify the determination of object state change.

\noindent\textbf{The final answer from temporal cross-view reasoning.} 
We then derive the final answer from the intermediate $k$ answers.  When $k$ intermediate answers are consistent, \name{} adopts the consensus. When intermediate answers are inconsistent, \name{} reconciles them by jointly accounting for viewpoint variations and leveraging the chronological order of the frames. This enables the prioritization of the most informative observations while reasoning about temporal progression.

 \emph{(1) Cross-view reasoning.} Since the retrieved frames are captured from different viewpoints and at different times, they vary in how informative they are about the object's state. Factors such as occlusion or camera angle may cause some frames to miss the object even when it is present, leading to misleading conclusions. To address this, we include 
 inputs to the MLLM that explicitly guide the model in reconciling such inconsistencies through cross-validation across multiple frames. We also add few-shot exemplars. When intermediate answers disagree with each other, the model evaluates which frames offer more reliable visual cues. For example, an exemplar instructs that if most frames show no target object but one frame clearly captures it, it can be inferred that the missing object in other frames may be due to poor viewing angles or occlusions, rather than the conclusion that the object never existed. By recognizing and attributing discrepancies to uninformative viewpoints, the model is better equipped to detect true state changes. 

 \emph{(2) Temporal progress-based reasoning.} An object may appear in earlier frames and change its state in later ones, meaning that temporal changes themselves can lead to inconsistent intermediate answers. Suppose the retrieved frames $\{I_{t_j}\}_{j=1}^{k}$ are captured at ordered timestamps \((t_1, t_2, \dots, t_k)\). We feed these frames to the MLLM in their chronological order, allowing it to reason with respect to temporal progression. We include %few-shot exemplars 
 instructions (including few-shot exemplars) that explicitly demonstrate how to analyze object presence across ordered frames. For instance, when an object exhibits a consistent presence in earlier timestamps but then disappears in later ones, this temporal trajectory provides strong evidence for an actual disappearance event. Rather than treating such inconsistencies as noise, our approach leverages them to identify genuine state changes and infer that the object has disappeared.

\section{Experimental Setup}
\subsection{Baselines}
\label{sec:baselines}

To evaluate the effectiveness of retrieving relevant egocentric frames, we compare ObjChangeVR with 
Caption-CLIP, Image-CLIP, and Viewpoint-Retrieval.  
To examine the impact of reasoning strategies, we further compare \name{} against CoT-SC~\cite{wang2023self} adapted to our task, as well as \name{} without the temporal cross-viewpoint CoT, i.e., \mbox{\name{} w/o TCV}.

\noindent\textbf{Caption-CLIP.} Each frame is captioned using an MLLM (GPT-4o). 
Similar to the relevant frame retrieval method used in~\citet{tang-etal-2024-m3d},
we encode the question and the captions of all previous frames using CLIP~\cite{radford2021learning}, and rank the captions by their cosine similarity to the question in the embedding space. The MLLM is then prompted with textual input consisting of the question, the caption of the current frame, and the top-$k$ most relevant captions from previous frames.

\noindent\textbf{Image-CLIP.}
Similar to the relevant frame retrieval method in ~\citet{Liang2023}, 
we embed the current frame and all previous frames with CLIP and rank the previous frames by their cosine similarity to the current frame in the visual embedding space. The MLLM is then prompted with multimodal input consisting of the question, the current frame, and the $k$ most similar retrieved frames, as in our \name{}.

\noindent\textbf{Viewpoint-Retrieval.}
Previous frames are ranked based solely on the difference between their camera viewpoints and the current camera viewpoint.
Formally, the similarity score for a previous frame $i$ with the current frame is defined as
$
\text{sim}(i) = w_p \cdot d_{\text{pos}}(i) + w_o \cdot d_{\text{ornt}}(i),
$
where $d_{\text{pos}}(i)$ denotes the Euclidean distance between the frame $i$'s viewpoint position and the current viewpoint position, and $d_{\text{ornt}}(i)$ represents the angular difference between their orientations. 
We empirically set $w_p = w_o = 1$. 
As in our \name{}, the MLLM is then prompted with multimodal input.

\noindent\textbf{CoT-SC} \cite{wang2023self}. Adapting CoT-SC to our task, we conduct $S$ independent inference runs on the same inputs: the question, current image, and the $k$ retrieved historical images. 
The retrieval pipeline is the same as that of \name{}. We generate outputs using a softmax temperature of $t = 0.7$ and shuffle the order of retrieved frames to encourage diversity.
The final answer is obtained by aggregating the $S$ results via majority vote.
For fair comparison, we set $S=3$ to match the three reasoning steps used in \name{}.

\subsection{Evaluation Metrics}

\noindent\textbf{Exact match (EM)@$\tau$.} 
We extract the judgment clause (e.g.,  the clause that states whether the object disappeared) 
from each answer and compute a similarity score against the ground truth using normalized string matching. 
Strict EM, the proportion of predictions that exactly match the ground-truth answers at the string level, remains close to zero in our task. 
This is because the generated answers 
include explanatory text beyond the concise ground-truth labels (e.g., disappeared, always here). To address this, we count a prediction as correct if its similarity to the ground truth exceeds a threshold $\tau$. 
This relaxed metric, denoted as EM@$\tau$, captures near-exact correctness while tolerating  
wording differences. 
We set $\tau = 0.80$, informed by the sensitivity analysis showing that our results are insensitive to the choice of $\tau$ in $\{0.70, 0.75, 0.80, 0.85, 0.90\}$ (Table~\ref{tab:tau_sensitivity} in Appendix~\ref{sec:sensitivity}).

\noindent\textbf{Macro-F1.} To evaluate categorical accuracy, both the generated answer and the ground-truth answer are classified into one of three predefined semantic classes (disappeared, never there, or always been there) based on 
indicative phrases or keywords in their respective judgment clauses. We compare the generated and the ground-truth answers using EM@$\tau$. The macro-F1 score is computed by calculating the F1 score for each class independently and averaging them equally.

\noindent\textbf{Weighted-F1.} 
We report weighted-F1, which averages the per-class F1 scores weighted by the number of samples per class. 
It reflects overall performance under the imbalanced class distribution. 

\subsection{Hyperparameters and Default Settings}

By default, we set $k=3$.
For retrieving relevant past frames in Section~\ref{sec:Relevant Cross-view Frame Retrieval}, we set $\alpha=2$ and $\beta=2$, %with clipped ranges 
$(\mathrm{min}_o,\mathrm{cap}_o)=(7,30)$, and $(\mathrm{min}_p,\mathrm{cap}_p)=(30,80)$. GPT-4o is used as the default MLLM.

\begin{table*}[t]
\centering
\small
\setlength{\tabcolsep}{0.9pt}
\renewcommand{\arraystretch}{1.05}
\begin{tabularx}{\textwidth}{
>{\arraybackslash}p{2.3cm}
| >{\arraybackslash}p{3.1cm}
| >{\centering\arraybackslash}p{1cm} >{\centering\arraybackslash}p{1cm} >{\centering\arraybackslash}p{1.3cm}
| >{\centering\arraybackslash}p{1cm} >{\centering\arraybackslash}p{1cm} >{\centering\arraybackslash}p{1.3cm}
| >{\centering\arraybackslash}p{1cm} >{\centering\arraybackslash}p{1cm} >{\centering\arraybackslash}p{1.3cm}}
\toprule
\makecell[c]{\multirow{3}{*}{Model}} & \makecell[c]{\multirow{3}{*}{Method}}
& \multicolumn{3}{c|}{Short traj.}
& \multicolumn{3}{c|}{Long traj.}
& \multicolumn{3}{c}{All traj.} \\
\cmidrule(lr){3-5} \cmidrule(lr){6-8} \cmidrule(lr){9-11}
& & \makecell{EM\\@0.8} & \makecell{Macro\\F1} & \makecell{Weighted\\F1}
  & \makecell{EM\\@0.8} & \makecell{Macro\\F1} & \makecell{Weighted\\F1}
  & \makecell{EM\\@0.8} & \makecell{Macro\\F1} & \makecell{Weighted\\F1} \\
\midrule

\multirow{4}{*}{GPT-4o}
& Caption-CLIP        & 0.529 & 0.595 & 0.588 & 0.528 & 0.522 & 0.537 & 0.529 & 0.574 & 0.570 \\
& Image-CLIP          & 0.616 & 0.649 & 0.644 & 0.558 & 0.572 & 0.585 & 0.592 & 0.619 & 0.620 \\
& Viewpoint-Retrieval & 0.623 & 0.659 & 0.657 & 0.570 & 0.584 & 0.599 & 0.601 & 0.631 & 0.635 \\
& \name{}             & \textbf{0.822} & \textbf{0.830} & \textbf{0.837} & \textbf{0.652} & \textbf{0.661} & \textbf{0.669} & \textbf{0.754} & \textbf{0.770} & \textbf{0.774} \\
\midrule

\multirow{4}{*}{GPT-4o mini}
& Caption-CLIP        & 0.416 & 0.395 & 0.357 & 0.537 & 0.449 & 0.468 & 0.464 & 0.420 & 0.400 \\
& Image-CLIP          & 0.472 & 0.430 & 0.381 & 0.513 & 0.459 & 0.479 & 0.489 & 0.441 & 0.420 \\
& Viewpoint-Retrieval & 0.472 & 0.435 & 0.388 & 0.527 & 0.471 & 0.491 & 0.494 & 0.449 & 0.429 \\
& \name{}             & \textbf{0.696} & \textbf{0.692} & \textbf{0.699} & \textbf{0.589} & \textbf{0.573} & \textbf{0.579} & \textbf{0.653} & \textbf{0.656} & \textbf{0.657} \\
\midrule

\multirow{4}{*}{Gemini 2.0 Flash}
& Caption-CLIP        & 0.478 & 0.493 & 0.453 & 0.440 & 0.430 & 0.453 & 0.462 & 0.471 & 0.455 \\
& Image-CLIP          & 0.643 & 0.667 & 0.661 & 0.563 & 0.571 & 0.582 & 0.611 & 0.630 & 0.630 \\
& Viewpoint-Retrieval & 0.653 & 0.677 & 0.672 & 0.590 & 0.594 & 0.605 & 0.628 & 0.645 & 0.645 \\
& \name{}             & \textbf{0.786} & \textbf{0.806} & \textbf{0.811} & \textbf{0.604} & \textbf{0.615} & \textbf{0.624} & \textbf{0.713} & \textbf{0.739} & \textbf{0.741} \\
\bottomrule
\end{tabularx}
%\vspace{-0.2cm}
\caption{Comparison of \name{} and 
different relevant frame retrieval methods 
across three MLLMs.}
%\vspace{-0.2cm}
\label{tab:retrieval_baselines}
\end{table*}

\begin{table*}[t]
\centering
\small
\setlength{\tabcolsep}{0.9pt}
\renewcommand{\arraystretch}{1.05}
\begin{tabularx}{\textwidth}{
>{\arraybackslash}p{2.3cm}
| >{\arraybackslash}p{3.1cm}
| >{\centering\arraybackslash}p{1cm} >{\centering\arraybackslash}p{1cm} >{\centering\arraybackslash}p{1.3cm}
| >{\centering\arraybackslash}p{1cm} >{\centering\arraybackslash}p{1cm} >{\centering\arraybackslash}p{1.3cm}
| >{\centering\arraybackslash}p{1cm} >{\centering\arraybackslash}p{1cm} >{\centering\arraybackslash}p{1.3cm}}
\toprule
\makecell[c]{\multirow{3}{*}{Model}} & \makecell[c]{\multirow{3}{*}{Method}}
& \multicolumn{3}{c|}{Short traj.}
& \multicolumn{3}{c|}{Long traj.}
& \multicolumn{3}{c}{All traj.} \\
\cmidrule(lr){3-5} \cmidrule(lr){6-8} \cmidrule(lr){9-11}
& & \makecell{EM\\@0.8} & \makecell{Macro\\F1} & \makecell{Weighted\\F1}
  & \makecell{EM\\@0.8} & \makecell{Macro\\F1} & \makecell{Weighted\\F1}
  & \makecell{EM\\@0.8} & \makecell{Macro\\F1} & \makecell{Weighted\\F1} \\
\midrule

\multirow{3}{*}{GPT-4o}
& CoT-SC            & 0.754 & 0.745 & 0.754 & 0.607 & 0.623 & 0.631 & 0.695 & 0.702 & 0.708 \\
& \name{} w/o TCV   & 0.745 & 0.737 & 0.747 & 0.590 & 0.611 & 0.620 & 0.683 & 0.699 & 0.700 \\
& \name{}           & \textbf{0.822} & \textbf{0.830} & \textbf{0.837} & \textbf{0.652} & \textbf{0.661} & \textbf{0.669} & \textbf{0.754} & \textbf{0.770} & \textbf{0.774} \\
\midrule

\multirow{3}{*}{GPT-4o mini}
& CoT-SC            & 0.537 & 0.536 & 0.518 & 0.524 & 0.536 & 0.549 & 0.536 & 0.545 & 0.532 \\
& \name{} w/o TCV   & 0.527 & 0.528 & 0.511 & 0.536 & 0.536 & 0.549 & 0.531 & 0.541 & 0.529 \\
& \name{}           & \textbf{0.696} & \textbf{0.692} & \textbf{0.699} & \textbf{0.589} & \textbf{0.573} & \textbf{0.579} & \textbf{0.653} & \textbf{0.656} & \textbf{0.657} \\
\midrule

\multirow{3}{*}{Gemini 2.0 Flash}
& CoT-SC            & 0.688 & 0.703 & 0.709 & 0.541 & 0.559 & 0.565 & 0.629 & 0.649 & 0.653 \\
& \name{} w/o TCV   & 0.669 & 0.682 & 0.689 & 0.533 & 0.556 & 0.562 & 0.615 & 0.636 & 0.639 \\
& \name{}           & \textbf{0.786} & \textbf{0.806} & \textbf{0.811} & \textbf{0.604} & \textbf{0.615} & \textbf{0.624} & \textbf{0.713} & \textbf{0.739} & \textbf{0.741} \\
\bottomrule
\end{tabularx}
\caption{Comparison of \name{} and variants with different reasoning methods across three MLLMs.} 
%\vspace{-0.3cm}
\label{tab:reasoning_variants}
\end{table*}

\section{Experimental Results}

\noindent\textbf{Overall performance of \name{}.} Tables~\ref{tab:retrieval_baselines} and~\ref{tab:reasoning_variants} report EM@$0.8$, macro-F1, and weighted-F1 scores on both short and long video trajectories. Table~\ref{tab:retrieval_baselines} 
shows the comparison of \name{} against different relevant frame retrieval methods, while Table~\ref{tab:reasoning_variants} presents the comparison of ObjChangeVR against variants with different reasoning methods.
For the question-answering task on object state change, ObjChangeVR consistently outperforms all other methods across both types of video recordings on all metrics. In particular, using GPT-4o as the MLLM, \name{} achieves an EM@$0.8$ of 0.822 on short trajectories and 0.652 on long trajectories, resulting in an overall average EM@$0.8$ of 0.754. 
Interestingly, Viewpoint-Retrieval outperforms both Caption-CLIP and Image-CLIP across short and long video recordings, indicating that the position and orientation of viewpoints prove valuable for retrieving relevant frames in VR environments. Compared with  Caption-CLIP, Image-CLIP, and Viewpoint-Retrieval, the superior performance of \name{} suggests that 
it enables more effective retrieval of frames containing information to answer questions regarding object state change. The performance gains over CoT-SC and \name{} w/o TCV confirm the effectiveness of 
reasoning about the retrieved frames, reconciling conflicting cues across frames, and producing more 
accurate answers.

\noindent\textbf{Impact of MLLMs.} Tables~\ref{tab:retrieval_baselines} and~\ref{tab:reasoning_variants}  also show that \name{} consistently outperforms all other methods with different MLLMs (GPT-4o, GPT-4o mini, and Gemini 2.0 Flash), demonstrating its effectiveness regardless of model size or architecture. The performance gains over other methods vary across models: for instance, with GPT-4o, ObjChangeVR outperforms CoT-SC by 5.9\% in EM@0.8, while with GPT-4o mini and Gemini 2.0 Flash, the improvements are 11.7\% and 8.4\%, respectively. Smaller models benefit more from our approach, suggesting that \name{}'s retrieval and reasoning framework helps compensate for the performance gap in smaller-scale models.

\begin{table}[t]
  \centering
  \setlength{\tabcolsep}{6pt} 
  \small 
  \begin{tabular}{l l c c c c}
    \toprule
    Metric & Case &CoT-SC  & \name{} \\
    \midrule
    % ---------- EM@0.8 ----------
    \multirow{3}{*}{EM@0.8} 
      & Cons.   &0.748  &0.795  \\
      & Incons. &0.637  &0.709  \\
      & All &0.695 &0.754 \\
      \hline
    \addlinespace[2pt]
    % ---------- Macro-F1 ----------
    \multirow{3}{*}{Macro-F1} 
      & Cons.   &0.520  &0.749  \\
      & Incons. &0.551  &0.624  \\
      & All  &0.702 &0.770 \\
       \hline
    \addlinespace[2pt]
    % ---------- Weighted-F1 ----------
    \multirow{3}{*}{Weighted-F1} 
      & Cons.   &0.700  &0.810  \\
      & Incons. &0.664  &0.720  \\
      & All  &0.708  &0.774 \\
    \bottomrule
  \end{tabular}
  \caption{Performance of \name{} under consistent and inconsistent intermediate answers.}
  \label{tab:consistent}
    %\vspace{-0.4cm}
\end{table}

\begin{table}[t]
  \centering
  \setlength{\tabcolsep}{1.6pt}
  \small
  \begin{tabular}{@{}c l c c c c c c c@{}}
    \toprule
    \multirow{2}{*}{\makecell{Metric}} &
    \multirow{2}{*}{\makecell{Traj.}} &
    \multicolumn{6}{c}{$k$} \\
    \cmidrule(lr){3-8}
    & & 1 & 2 & 3 & 5 & 7 & 9 \\
    \midrule
    \multirow{3}{*}{\makecell{EM\\@0.8}} 
      & Short   &0.691 &0.774&0.822&0.776&0.711&0.672 \\
      & Long & 0.542& 0.638&0.652&0.598& 0.554&0.532\\
      & All &0.631 &0.720 &0.754 &0.705 &0.648 &0.616\\
       \hline
    \addlinespace[2pt]
    \multirow{3}{*}{\makecell{Macro-\\F1}} 
      & Short   & 0.787& 0.832& 0.830&0.823& 0.812&0.801 \\
      & Long & 0.612& 0.656&  0.661&  0.656& 0.648&  0.651 \\
       & All &0.727 &0.770 &0.770 &0.762 &0.751 &0.743\\
        \hline
    \addlinespace[2pt]
    \multirow{3}{*}{\makecell{Weighted-\\F1}} 
      & Short   & 0.794& 0.838&0.837&0.831& 0.824&0.812 \\
      & Long &0.621&0.664& 0.669&0.662&0.654& 0.655 \\
       & All &0.730 &0.773 &0.774 &0.767 &0.758 &0.750\\
    \bottomrule
  \end{tabular}
  \caption{Performance of \name{} across different numbers of retrieved frames ($k$).}
  \label{tab:topk}
  %\vspace{-0.4cm}
\end{table}

\begin{table}[t]
\centering
\small
\begin{tabular}{lcc}
\hline
Variant & Zero-shot & Few-shot \\
\hline
EM              & 0.731              & 0.754             \\
Macro-F1        & 0.734              & 0.770             \\
Weighted-F1     & 0.740              & 0.774             \\
\hline
\end{tabular}
\caption{Comparison between the ObjChangeVR variants with zero-shot and few-shot prompting.}
%\vspace{-0.4cm}
\label{tab:zeroshot_fewshot}
\end{table}

\noindent\textbf{Performance under consistent and inconsistent intermediate answers.} We evaluate the reasoning capability under consistent and inconsistent intermediate answers. 
We test both CoT-SC and ObjChangeVR and report results in~Table~\ref{tab:consistent}, as these are two methods that use intermediate answers. CoT-SC produces inconsistent intermediate reasoning answers in 47.9\% of the questions. In comparison, ObjChangeVR reduces this inconsistency ratio to 33.2\%, demonstrating an improvement in producing consistent reasoning paths. Across all metrics, \name{} outperforms CoT-SC under both consistent and inconsistent intermediate answers. In particular, under inconsistent intermediate answers, ObjChangeVR achieves 7.2\%,  7.3\%, and 5.6\% improvements in EM@0.8, macro-F1, and weighted-F1 over CoT-SC, respectively. These results demonstrate that ObjChangeVR’s temporal cross-view reasoning enhances robustness by generating reliable final answers even with varying reasoning paths.

\noindent\textbf{Impact of $k$.}
We examine the impact of the number of retrieved frames ($k$) in Table~\ref{tab:topk}. For different $k$, \name{} consistently 
performs better on short trajectories than on long ones, suggesting that it is more effective when the temporal spans and viewpoint changes are smaller along the trajectory. 
As $k$ increases from 1 to 3, we observe consistent improvements in EM@0.8, macro-F1, and weighted-F1 scores. This indicates that retrieving multiple frames provides richer contextual information compared to relying on a single frame, 
where additional frames 
capture objects from different viewpoints or time periods.

As $k$ increases from 3 to 9, we observe a 
performance decline. For instance, EM@0.8 decreases by 15.0\% on short trajectories and 12.0\% on long ones. To investigate the reason behind this trend,  we analyze how intermediate reasoning inconsistency affects final answers.
Figure~\ref{fig:consistent_inconsistent} in Appendix~\ref{sec:kdetail} shows a drop in the proportion of consistent intermediate answers (with a 31.0\% drop from $k=3$ to $k=9$), indicating that retrieving more frames increases the chance of introducing conflicting or distracting contextual information
as $k$ grows too large. 
This rise in inconsistent reasoning 
contributes to the performance decline observed at larger $k$. 
Based on the observation, we select $k=3$ as the default setting in the paper. Retrieving a small number of previous frames (e.g., 3) achieves the best performance by providing richer contextual information than a single frame, while reducing the negative impacts brought by inconsistent intermediate reasoning for larger $k$. It also reduces token consumption and inference latency.

\noindent\textbf{Impact of $\tau$.}
We examine whether our evaluation results are sensitive to the choice of the similarity threshold $\tau$ used in EM@$\tau$. 
Specifically, we evaluate all methods with $\tau \in \{0.70, 0.75, 0.80, 0.85, 0.90\}$. As shown in Table~\ref{tab:tau_sensitivity} in Appendix~\ref{sec:sensitivity}, the relative ranking of all methods remains unchanged across different $\tau$, and ObjChangeVR consistently achieves the best performance. Moreover, the absolute variations in EM@$\tau$ across different $\tau$ are small, indicating that our conclusions are %robust %and 
not sensitive
to the specific choice of $\tau$.

\noindent\textbf{Impact of few-shot prompting.}
\label{sec:fewshot}
To better understand the role of few-shot prompting in \name{}, we evaluate a zero-shot variant. 
In this variant, we remove all few-shot exemplars from both the pairwise comparison prompt and the final reasoning prompt. The prompts only describe how to reason (e.g., using the temporal ordering of retrieved frames and the spatial similarity of viewpoints), without providing any exemplars. 
Table~\ref{tab:zeroshot_fewshot} reports the comparison between the ObjChangeVR variants with zero-shot and few-shot prompting. It shows that removing all exemplars leads to only a small performance decrease, and ObjChangeVR with zero-shot prompting still outperforms all other methods (the performance of other methods is listed in Tables~\ref{tab:retrieval_baselines} and~\ref{tab:reasoning_variants}). These findings suggest that while few-shot prompting provides a modest benefit, the effectiveness of our reasoning module does not heavily rely on it.

\section{Conclusion}
In this paper, we introduce the \name{}-Dataset for the question-answering task of object change detection with continuous egocentric views. We also propose \name{} to effectively retrieve relevant frames that contain useful information to answer the object state change query, and then use cross-view reasoning and temporal progress-based reasoning to get a final answer. 
Experimental results demonstrate that \name{} achieves higher reasoning accuracy compared with various methods and variants across both short and long trajectories and under consistent and inconsistent intermediate reasoning answers.

\section*{Acknowledgment}
We thank the anonymous reviewers for their constructive comments. This work was supported in part by NSF grant No. 2550742.

\section*{Limitations}
Our study has several limitations. 
First, due to limited computational resources, we were unable to deploy  MLLMs capable of processing multiple retrieved images in a single prompt on local servers or workstations. Hence, we could not evaluate several representative MLLMs locally. Second,  we primarily focus on object state changes where an object disappears. 
We focus on object disappearance as it represents a particularly challenging scenario: it requires reasoning about what may be no longer visible in the current query image. 
Other change types (e.g., object additions or movements) are yet to be explored. To provide preliminary insights, we collected a small-scale dataset on object additions and evaluated our method on it in Appendix~\ref{sec:object_addition}. 
Third, our data collection process requires manual trajectory sampling and partial human verification, limiting our ability to scale the dataset.

\section*{Ethical Considerations}

All data collection in this study was conducted entirely within virtual environments. User trajectories were generated through manual navigation in VR, thereby eliminating the risk of human injury. This study does not involve 
human subjects, and the collected data does not involve personal or sensitive information. All VR scenes used for data generation were obtained through purchase 
and
are permitted for academic research. Furthermore, our use of MLLMs adheres to applicable laws, licensing terms, and institutional guidelines.

\bibliography{custom}

\appendix

\section{Appendix}
\subsection{Details of \name{}-Dataset}
\label{sec:datasetdetail}
Scene scales and sections in \name{}-Dataset are listed in Table~\ref{tab:scene_stats}.

\begin{table*}[t]
\centering
\small
\begin{tabular}{>{\centering\arraybackslash}m{2.8cm}
                >{\centering\arraybackslash}m{1.6cm}
                p{8cm}}
\hline
Scene & Scale & \multicolumn{1}{c}{Scene section} \\
\hline
Villa interior & Small &
First-floor kitchen, first-floor dining room, first-floor living room, first-floor balcony, first-floor lounge, first-floor storage room, second-floor bathroom, second-floor master bedroom with bed and wardrobe, second-floor study room with bookshelf and desk \\
\hline
Restaurant & Small &
Dining area, men's restroom, women's restroom, entrance, checkout counter, pickup area, bar counter \\
\hline
Market & Small &
Fish shop, dessert shop, butcher shop, weapon shop, vegetable stall \\
\hline
Museum & Large &
Ground floor main exhibition hall, ground floor small exhibition room, ground floor corridor exhibition area, mezzanine staircase area, second floor corridor exhibition area, second floor balcony \\
\hline
Viking village & Large &
Large house, small house, barrel storage, blacksmith workshop, dock, shipbuilding area, gate and tower, fenced area with multiple items \\
\hline
\end{tabular}
\caption{Scene sections in \name{}-Dataset.}
\label{tab:scene_stats}
\end{table*}

\subsection{Comparison of \name{}-Dataset and Existing 3D and Video-based Benchmarks}
\label{sec:dataset-comparision}
We compare the \name{}-Dataset with representative 3D and video-based benchmarks across multiple dimensions. The comparison is summarized in Table~\ref{tab:dataset_comparison}.

\begin{table*}[t]
\centering
\scriptsize
\renewcommand{\arraystretch}{1.15}
\setlength{\tabcolsep}{0.4pt}

\begin{tabular}{
>{\centering\arraybackslash}p{4cm}  
>{\centering\arraybackslash}p{1.5cm}  
>{\centering\arraybackslash}p{1.8cm}  
>{\centering\arraybackslash}p{1.5cm}  
>{\centering\arraybackslash}p{1.8cm} 
>{\centering\arraybackslash}p{1.5cm}  
>{\centering\arraybackslash}p{1.5cm}  
>{\centering\arraybackslash}p{2.2cm}  
}

\toprule
\makecell[c]{\multirow{2}{*}{\centering Dataset}}  &
Virtual env.-based &
Natural language Q\&A &
\makecell[c]{\multirow{2}{*}{\centering Egocentric}}  &
Object state change &
\makecell[c]{\multirow{2}{*}{\centering Camera pose}} &
\makecell[c]{\multirow{2}{*}{\centering Avg. duration}}  &
\makecell[c]{\multirow{2}{*}{\centering Scene type}}\\
\midrule

ScanQA~\citep{ScanQA} & \xmark & \cmark & \xmark & \xmark & \cmark & N/A & Indoor \\
MMScan~\citep{MMScan} & \xmark & \cmark & \xmark & \xmark & \cmark & N/A & Indoor \\
VLA-3D~\citep{VLA-3D} & \cmark & \xmark & \xmark & \xmark & \xmark & N/A & Indoor \\
SIMMC-VR~\citep{wu2023simmc} & \cmark & \cmark & \cmark & \xmark & \cmark & $\sim$2 min & Indoor \\
ChangeSim~\citep{ChangeSim} & \cmark & \xmark & \xmark & \cmark & \cmark & N/A & Indoor \\
Causal-VidQA~\citep{Causal-VidQA} & \xmark & \cmark & \xmark & \cmark & \xmark & $>$9 s & Indoor/Outdoor \\
FunQA~\citep{FunQA} & \xmark & \cmark & \xmark & \cmark & \xmark & 19 s & Indoor/Outdoor \\
SurveillanceVQA~\citep{SurveillanceVQA-589K} & \xmark & \cmark & \xmark & \cmark & \xmark & N/A & Indoor/Outdoor \\
Pano-AVQA~\citep{Pano-AVQA} & \xmark & \cmark & \xmark & \cmark & \xmark & $\sim$5 s & Indoor/Outdoor \\
VideoEspresso~\citep{VideoEspresso} & \xmark & \cmark & \xmark & \cmark & \xmark & N/A & Indoor/Outdoor \\
Ego4D~\citep{Ego4D} & \xmark & \cmark & \cmark & \cmark & \xmark & $\sim$8 min & Indoor/Outdoor \\
EgoTracks~\citep{tang2023egotracks} & \xmark & \xmark & \cmark & \cmark & \cmark & $\sim$6 min & Indoor/Outdoor \\
EPIC-KITCHENS~\citep{damen2021rescaling} & \xmark & \xmark & \cmark & \cmark & \cmark & $\sim$7.5 min & Indoor \\
\midrule
 \name{}-Dataset (\emph{ours})& \cmark & \cmark & \cmark & \cmark& \cmark & $\sim$60/180 s & Indoor/Outdoor \\

\bottomrule
\end{tabular}
\caption{Comparison of the \name{}-Dataset with representative 3D and video-based benchmarks.}
\label{tab:dataset_comparison}
\end{table*}

\subsection{Impact of the Similarity Threshold $\tau$}
\label{sec:sensitivity}
Table~\ref{tab:tau_sensitivity} shows the impact of the threshold $\tau$ on the performance of different methods.

\subsection{Influence of $k$ on the Intermediate Answers}
\label{sec:kdetail}
\begin{figure}[t]
    \centering
    \includegraphics[width=0.8\columnwidth]{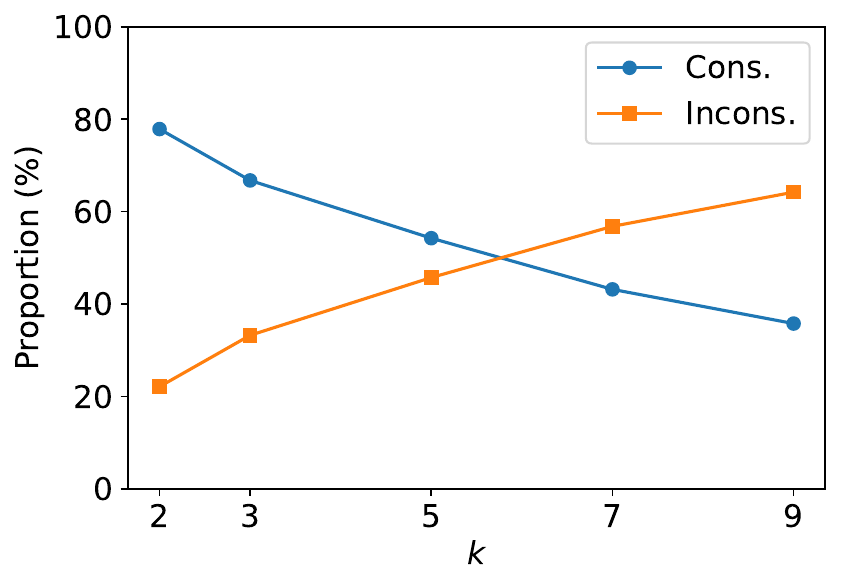}
    \caption{Proportion of questions (out of 5,000) with consistent and inconsistent intermediate answers across different $k$.}
    \label{fig:consistent_inconsistent}
\end{figure}

The number of consistent and inconsistent intermediate reasoning answers under different $k$ is shown in Figure~\ref{fig:consistent_inconsistent}. As $k$ increases, the proportion of consistent intermediate answers gradually decreases, while the proportion of inconsistent answers rises.

\subsection{Object Addition}
\label{sec:object_addition}
Our pipeline can potentially be generalized to different change types (e.g., object additions and movements). 
To demonstrate this potential, we collected a small-scale dataset of 100 questions on object additions using our established pipeline and evaluated our method on this dataset. 
The results are shown in Table~\ref{tab:obj_addition_results}. 
In future work, we plan to expand this benchmark by constructing full-scale datasets for categories such as object movement and addition.

\subsection{Inference Latency}

We investigate the inference latency by measuring retrieval time, reasoning time, and total time for each method (Table~\ref{tab:latency}). We use GPT-4o as the MLLM. Our method incurs additional computation, mainly from the reasoning stage, since it requires processing multiple cross-view and cross-time frames. The total inference time remains under $10$s.

\subsection{Statistical Robustness of Evaluation Results}
To assess whether the observed performance differences are robust to dataset size, we conduct a 1,000-sample bootstrap resampling analysis over all question-answer pairs. The 95\% confidence intervals (CI95\%) for EM@0.8 across all methods are shown in Table~\ref{tab:ci95}. The tight intervals indicate low variance and show that the performance differences (especially the gains of ObjChangeVR) are statistically robust.

\begin{table*}[t]
\centering
\small
\begin{tabular}{lcccccc}
\hline
Method &
Caption-CLIP &
Image-CLIP &
Viewpoint-Retrieval &
CoT-SC&
ObjChangeVR w/o TCV &
ObjChangeVR \\
\hline
EM@0.70 & 0.5302 & 0.5934 & 0.6026 & 0.6958 & 0.6840 & \textbf{0.7588} \\
EM@0.75 & 0.5302 & 0.5934 & 0.6026 & 0.6958 & 0.6840 & \textbf{0.7566} \\
EM@0.80 & 0.5286 & 0.5924 & 0.6014 & 0.6950 & 0.6828 & \textbf{0.7540} \\
EM@0.85 & 0.5214 & 0.5860 & 0.5926 & 0.6774 & 0.6656 & \textbf{0.7368} \\
EM@0.90 & 0.5080 & 0.5678 & 0.5738 & 0.6626 & 0.6508 & \textbf{0.7182} \\
\hline
\end{tabular}
\caption{Sensitivity analysis of EM@$\tau$ under different similarity thresholds.}
\label{tab:tau_sensitivity}
\end{table*}

\begin{table*}[t]
\centering
\small
\begin{tabular}{lccc}
\toprule
Method & EM@0.8 & Macro-F1 & Weighted-F1 \\
\midrule
ObjChangeVR & 0.8200 & 0.8070 & 0.8307 \\
\bottomrule
\end{tabular}
\caption{Performance of ObjChangeVR on the object addition dataset.}
\label{tab:obj_addition_results}
\end{table*}

\begin{table*}[t]
\centering
\small
\begin{tabular}{lcccccc}
\hline
Method &
Caption-CLIP &
Image-CLIP &
Viewpoint-Retrieval &
CoT-SC &
ObjChangeVR w/o TCV &
ObjChangeVR \\
\hline
Retrieval (s) & 0.035 & 0.261 & 0.003 & 0.004 & 0.004 & 0.003 \\
Reasoning (s) & 2.383 & 3.372 & 3.194 & 8.249 & 2.814 & 9.562 \\
Total (s)     & 5.887 & 3.653 & 3.323 & 8.256 & 2.821 & 9.566 \\
\hline
\end{tabular}
\caption{Inference latency breakdown of different methods.}
\label{tab:latency}
\end{table*}

\begin{table*}[t]
\centering
\small
\setlength{\tabcolsep}{12pt}
\begin{tabular}{lc}
\hline
Method & CI95\% (EM@0.80) \\
\hline
Caption-CLIP           & [0.5140, 0.5424] \\
Image-CLIP             & [0.5788, 0.6064] \\
Viewpoint-Retrieval    & [0.5886, 0.6158] \\
CoT-SC                 & [0.6812, 0.7076] \\
ObjChangeVR w/o TCV    & [0.6696, 0.6950] \\
ObjChangeVR            & [0.7428, 0.7670] \\
\hline
\end{tabular}
\caption{95\% confidence intervals (CI95\%) of EM@0.80 obtained via 1,000-sample bootstrap resampling over all question--answer pairs.}
\label{tab:ci95}
\end{table*}

\subsection{Question-answer Pair Generation Prompt}
To automatically generate disappearance reasoning question–answer pairs, we provide the MLLMs with a prompt that specifies the temporal relationship between two frames of the same VR scene. The first image represents the \textit{previous} frame, while the second denotes the \textit{current} frame. The model is instructed to generate ten question–answer pairs by comparing the two frames and reasoning about the presence or disappearance of objects over time.

The prompt enforces several key rules to ensure reasoning consistency. Questions must use environmental spatial anchors (e.g., ``the chair near the window’’). 
Each set of ten questions contains four about disappearing objects, three about objects that are present across both images, and three about objects that never appear in the scene. This balanced composition is designed to ensure sample diversity and prevent category bias during dataset generation. When fewer than four disappearing objects exist, the model is encouraged to rephrase similar 
queries with varied sentence structures to enhance linguistic diversity.
The instruction provided to the model is shown in Table~\ref{tab:generateqa-prompt}. An example of the output is shown in Table~\ref{tab:generateqa-example}.

\begin{table*}[t]
\centering
\renewcommand{\arraystretch}{1.15}
\setlength{\tabcolsep}{6pt}

% -------------- BEGIN CONTENT --------------
\begin{tcolorbox}[colback=gray!3, colframe=black!20, boxrule=0.3pt, left=3pt, right=3pt, top=3pt, bottom=3pt, width=\textwidth]
\textbf{[Instruction]}

You are given two images of the same VR scene taken at different times. 

The first image: previous screenshot. 

The second image: current screenshot.

Your task:

Generate 10 simple question-answer pairs comparing the images, focusing on object disappearance or persistence.

Rules:

- Use spatial references (e.g., "the chair near the window"). The reference should be the environment such as the left and right walls, windows and arches, rather than specific objects.

- If an object was visible in the first image but is missing in the second image, answer: "It has disappeared."

- If an object was present in both, answer: "It has always been here."

- If an object was missing in the first image, answer: "It was never there."

- Each time, ask 4 disappearing objects, 3 always present objects, and 3 objects that have never been in the scene. (If there are fewer than four disappearing objects in the scene, you can repeatedly ask questions about the same disappearing object using different sentences.) But try as hard as possible to ask questions about different objects.

- Only questions about whether an object existed in the past are allowed. Only questions like "Did a specific object exist in a past scene?" are allowed. No other content can be asked. (When asking questions, never mention the two pictures and do not use words like "the differences or changes between the two pictures".) Use the past tense.

- When asking questions, the tone should seem as if you can only see the second picture (that is, the current picture), so when referring to relative positions, do not use objects that existed in the previous picture but not in the current one as references. When answering, you can see the information of the two pictures.

- Respect facts, not illusions.

Output Format:

1. <question>
   Answer: <answer>

2. <question>
   Answer: <answer>

\end{tcolorbox}

\caption{Instruction of question-answer pair generation.}
\label{tab:generateqa-prompt}
\end{table*}

\begin{table*}[t]
\centering
\renewcommand{\arraystretch}{1.15}
\setlength{\tabcolsep}{6pt}

\begin{tcolorbox}[colback=gray!3, colframe=black!20, boxrule=0.3pt, left=3pt, right=3pt, top=3pt, bottom=3pt, width=\textwidth]

\vspace{1em}
\textbf{[Input]}

\begin{center}

  \includegraphics[width=0.45\linewidth]{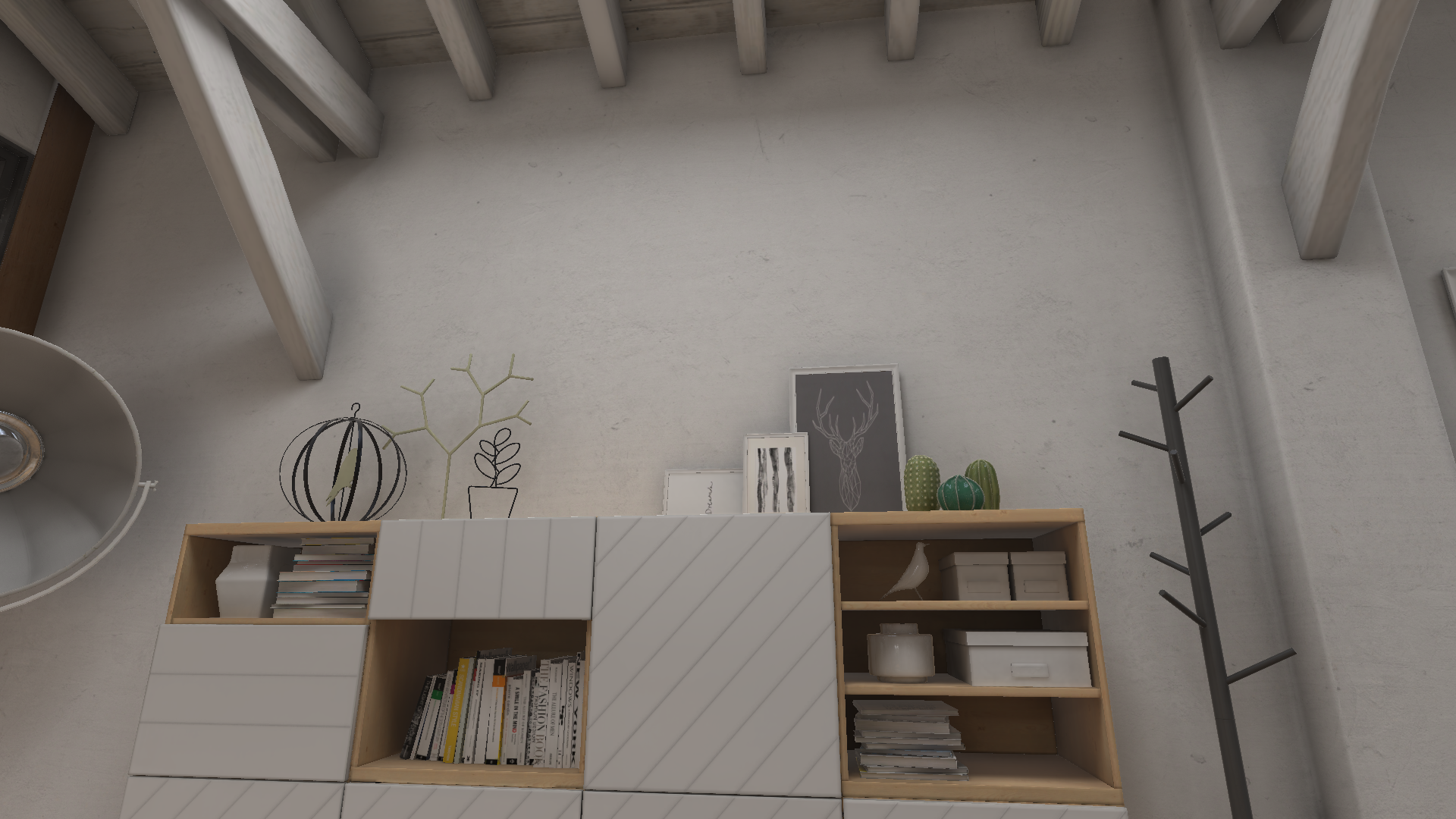}
  \hfill
  \includegraphics[width=0.45\linewidth]{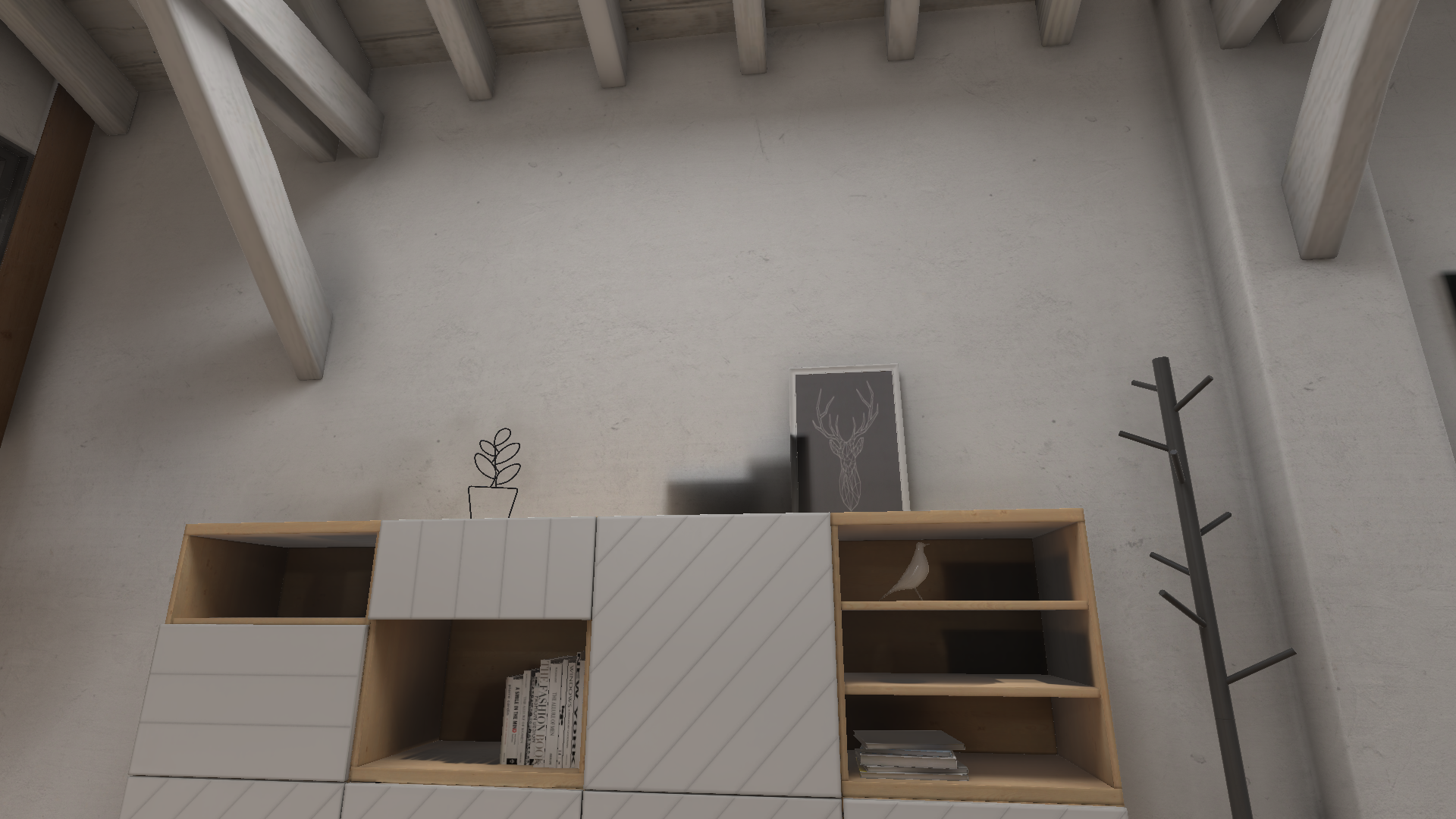}

\end{center}

\vspace{1em}
\textbf{[Output]}  

\textit{1. Was there a round wire plant decoration near the left wall in the past?}

Answer: It has disappeared.

\textit{2. Was there a small green cactus on the right side of the shelf in the past?}

Answer: It has disappeared.

\textit{3. Was there a second cactus next to the first one on the right side of the shelf in the past?}

Answer: It has disappeared.

\textit{4. Was there a thin white rectangular picture beside the deer painting near the right wall in the past?}

Answer: It has disappeared.

\textit{5. Was the framed deer picture near the right wall here before?}

Answer: It has always been here.

\textit{6. Was the small bird-shaped decoration on the right shelf here before?}

Answer: It has always been here.

\textit{7. Were the books stacked on the lower right shelf here before?}

Answer: It has always been here.

\textit{8. Was there a coat rack near the right wall in the past?}

Answer: It was never there.

\textit{9. Was there a lamp near the left side of the scene in the past?}

Answer: It was never there.

\textit{10. Was there a hanging decoration under the ceiling beams in the past?}

Answer: It was never there.
\end{tcolorbox}
% -------------- END CONTENT -------------- 

\caption{An example of question-answer pair generation.}
\label{tab:generateqa-example}
\end{table*}

\subsection{Case Study}
\label{sec:CaseStudy}
To clearly illustrate the reasoning process of \name{}, we present two representative cases: one with consistent intermediate reasoning and another with inconsistent intermediate reasoning. The consistent intermediate reasoning case is shown in Table~\ref{tab:casestudy1_consistent}, while the inconsistent one is shown in Table~\ref{tab:casestudy1_inconsistent}. These examples provide an intuitive understanding of how \name{} aggregates multiple intermediate reasoning answers and obtains the final decision.

\begin{table*}[t]
\centering
\renewcommand{\arraystretch}{1.15}
\setlength{\tabcolsep}{6pt}

% -------------- BEGIN CONTENT --------------
\begin{tcolorbox}[colback=gray!3, colframe=black!20, boxrule=0.3pt, left=3pt, right=3pt, top=3pt, bottom=3pt, width=\textwidth]

\vspace{1em}
\textbf{[Input]}

\texttt{<Question>}
\textit{Were there some food on the tables near the right wall?}

\texttt{<Current Frame>}

\begin{center}
  \includegraphics[width=0.4\linewidth]{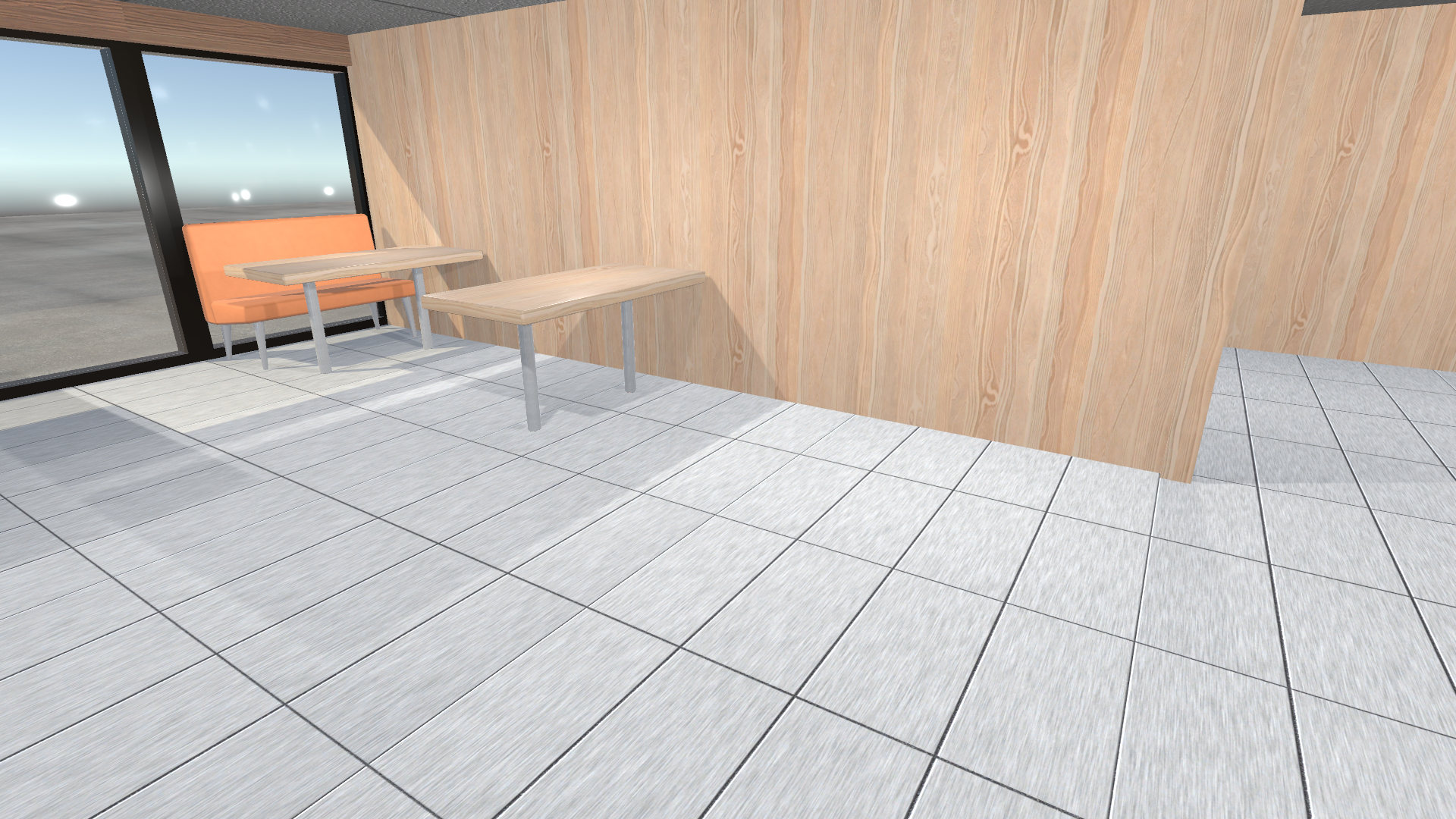}
  \vspace{2pt}
\end{center}

\texttt{<Retrieved Frames>}
\begin{center}
\begin{subtable}{0.32\linewidth}
  \centering
  \includegraphics[width=\linewidth]{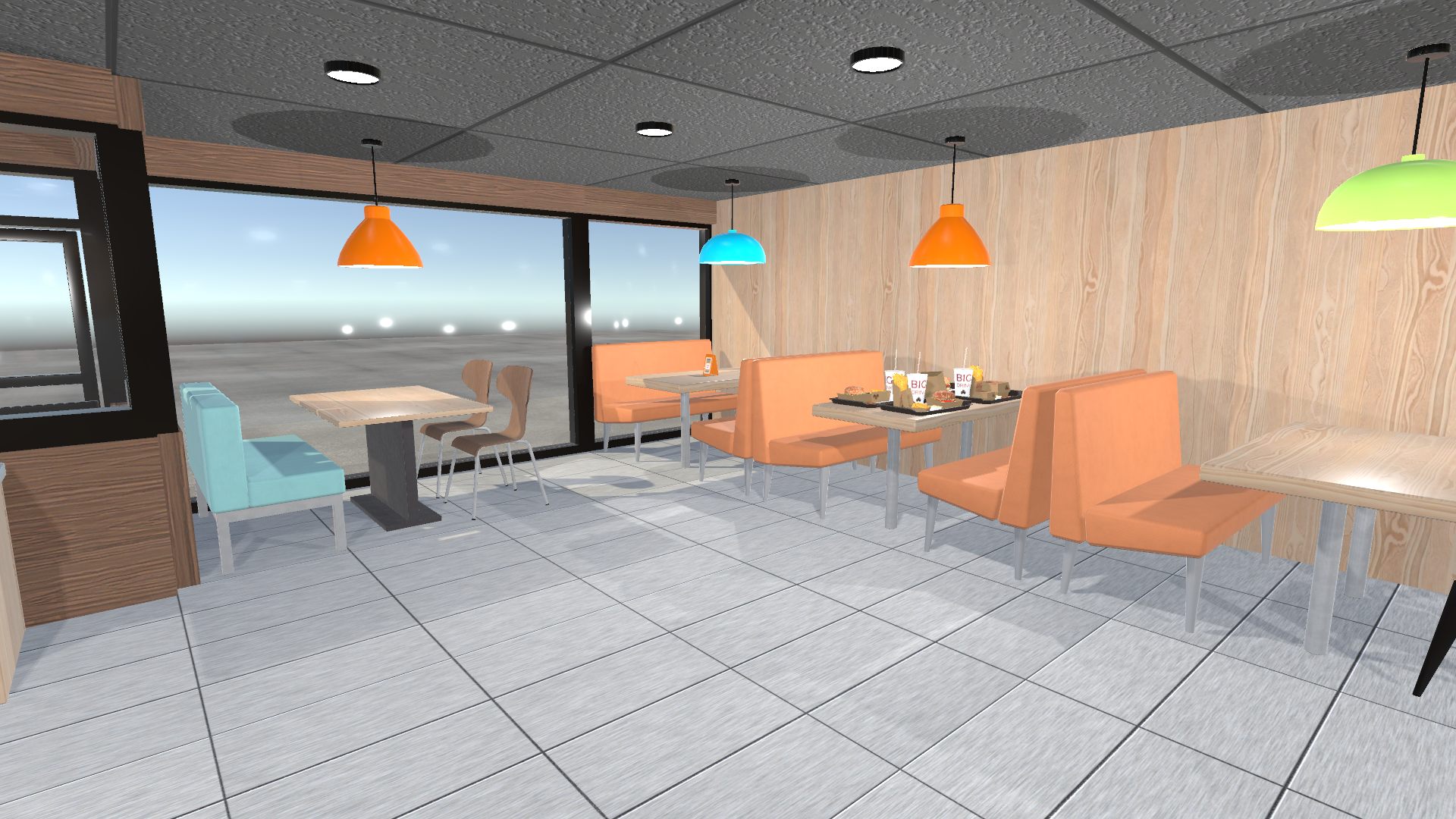}
  \caption{Frame 41}
\end{subtable}
\hfill
\begin{subtable}{0.32\linewidth}
  \centering
  \includegraphics[width=\linewidth]{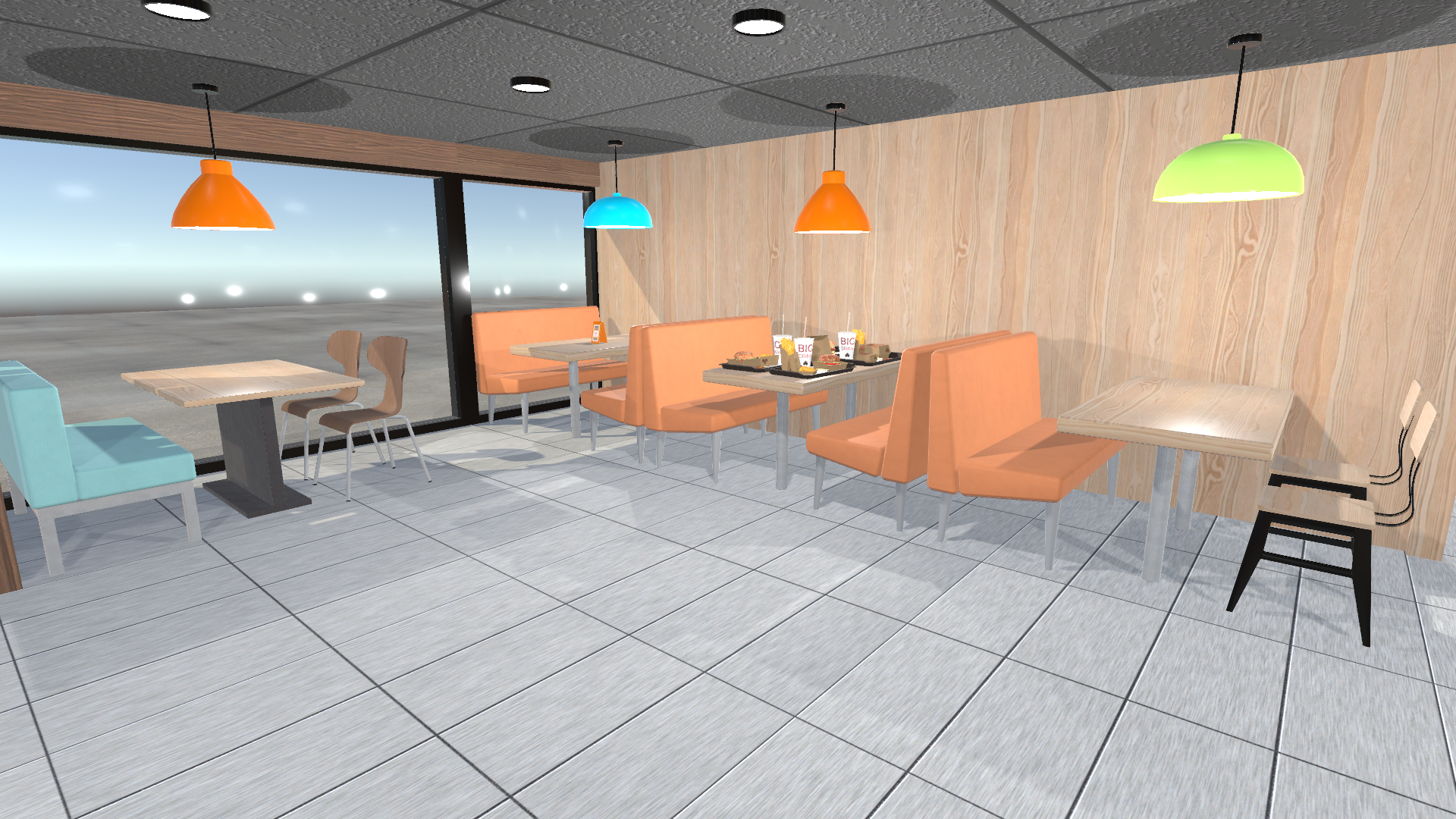}
  \caption{Frame 42}
\end{subtable}
\hfill
\begin{subtable}{0.32\linewidth}
  \centering
  \includegraphics[width=\linewidth]{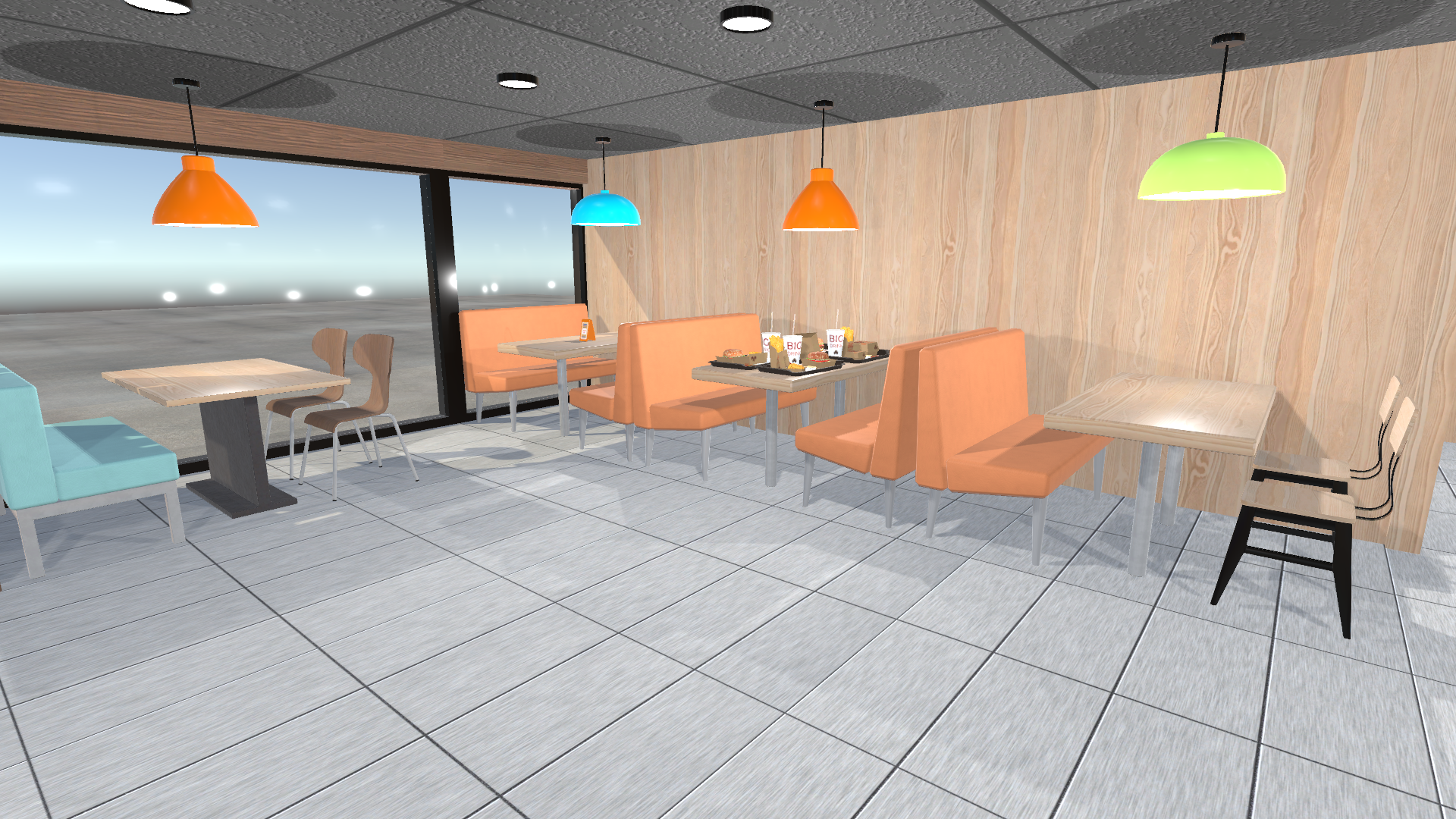}
  \caption{Frame 43}
\end{subtable}
\end{center}

\vspace{1em}
\textbf{[Output]}  

\texttt{<Intermediate Reasoning Answer 1>}
In the retrieved picture, you can see food items on the tables near the right wall, specifically on the tables with orange chairs. In the current picture, those tables near the right wall are empty, with no food items present. So it has disappeared.

\texttt{<Intermediate Reasoning Answer 2>}
 In the retrieved picture, there is food on the tables near the right wall. In the current picture, those tables are empty with no food present. So it has disappeared.

\texttt{<Intermediate Reasoning Answer 3>}
In the retrieved picture, there is food on the tables near the right wall. In the current picture, the tables near the right wall have no food on them. So it has disappeared.

\texttt{<Final Answer>}
It has disappeared. Because all three sub-answers consistently indicate that there was food on the tables near the right wall in the retrieved images, but the current image shows those tables as empty. This consistency across different retrieved images suggests that the food was indeed present before and has since disappeared.

\end{tcolorbox}

\caption{A case of consistent intermediate reasoning.}
\label{tab:casestudy1_consistent}
\end{table*}

\begin{table*}[t]
\centering
\renewcommand{\arraystretch}{1.15}
\setlength{\tabcolsep}{6pt}

\begin{tcolorbox}[colback=gray!3, colframe=black!20, boxrule=0.3pt, left=3pt, right=3pt, top=3pt, bottom=3pt, width=\textwidth]

\vspace{1em}
\textbf{[Input]}

\texttt{<Question>}
\textit{Was there a drink dispenser on the left counter?}

\texttt{<Current Frame>}

\begin{center}
  \includegraphics[width=0.4\linewidth]{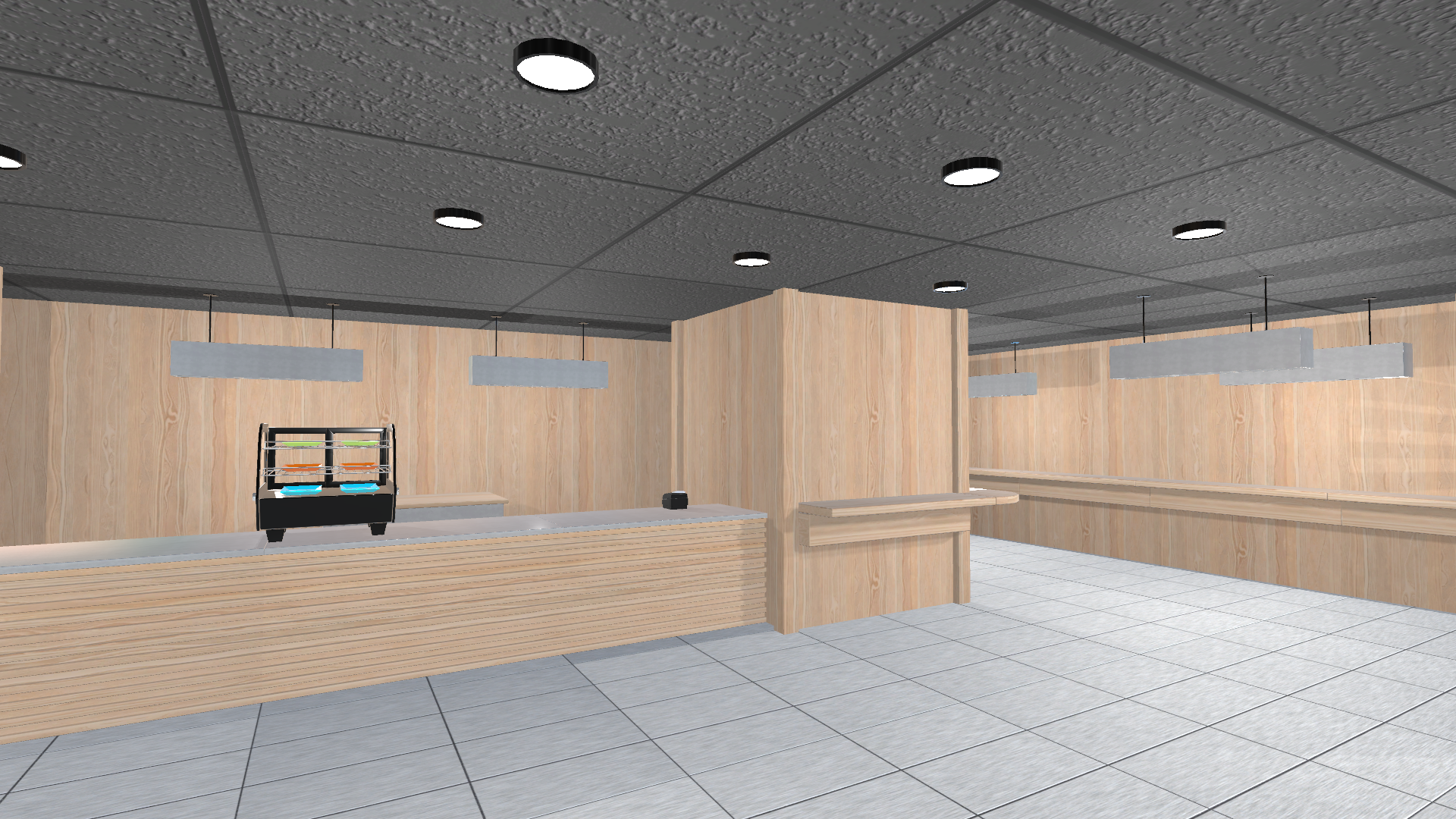}
  \vspace{2pt}
\end{center}

\texttt{<Retrieved Frames>}
\begin{center}
\begin{subtable}{0.32\linewidth}
  \centering
  \includegraphics[width=\linewidth]{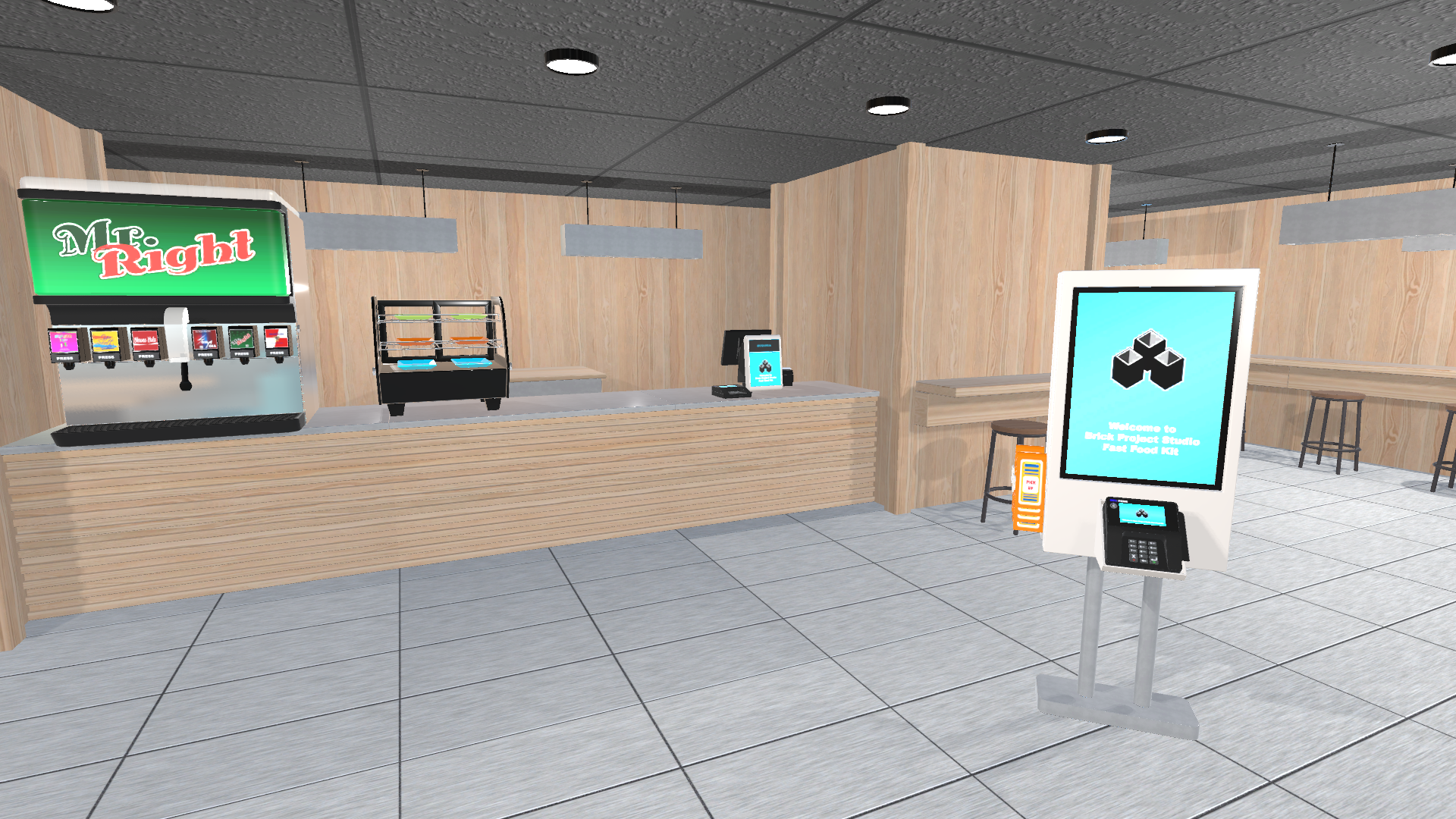}
  \caption{Frame 64}
\end{subtable}
\hfill
\begin{subtable}{0.32\linewidth}
  \centering
  \includegraphics[width=\linewidth]{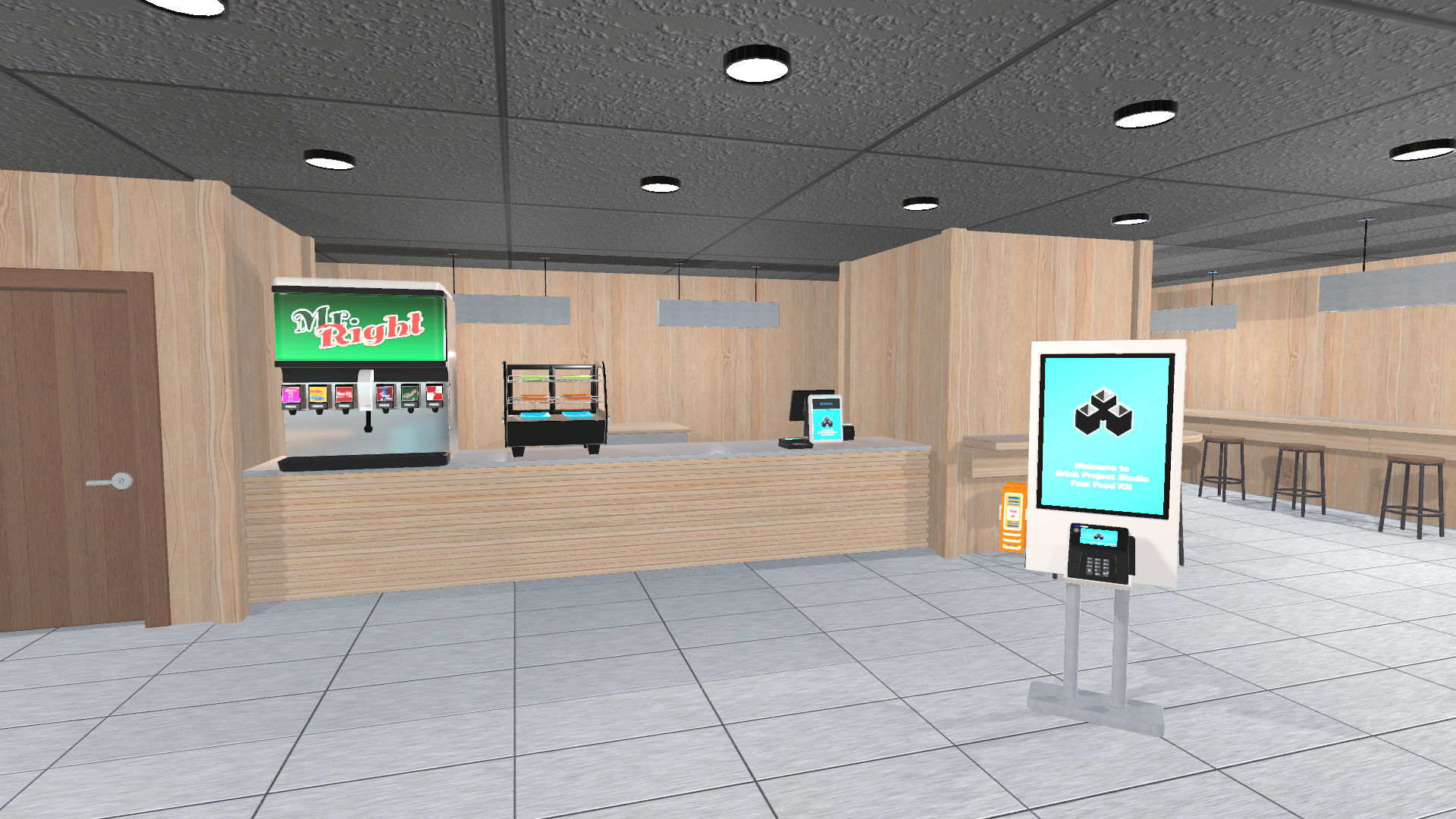}
  \caption{Frame 71}
\end{subtable}
\hfill
\begin{subtable}{0.32\linewidth}
  \centering
  \includegraphics[width=\linewidth]{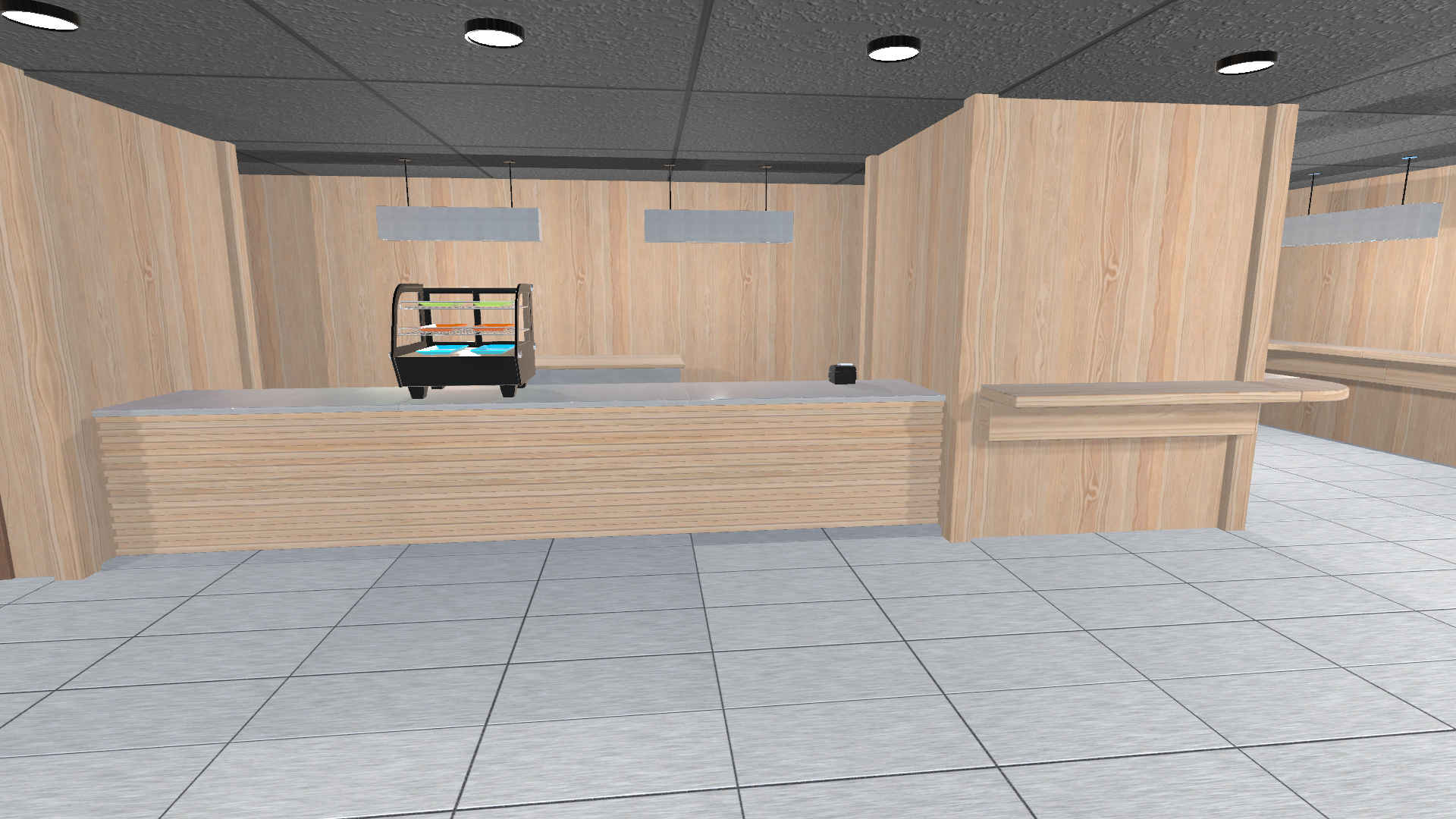}
  \caption{Frame 147}
\end{subtable}
\end{center}

\vspace{1em}
\textbf{[Output]}  

\texttt{<Intermediate Reasoning Answer 1>}
In the retrieved picture, you can clearly see a drink dispenser labeled "Mr. Right" on the left counter. In the current picture, that drink dispenser is no longer there. So it has disappeared.

\texttt{<Intermediate Reasoning Answer 2>}
In the retrieved picture, there is a drink dispenser on the left counter. In the current picture, that drink dispenser is no longer there. So it has disappeared.

\texttt{<Intermediate Reasoning Answer 3>}
In the retrieved picture, there is no drink dispenser visible on the left counter. Similarly, in the current picture, there is still no drink dispenser visible on the left counter. So it was never there.

\texttt{<Final Answer>}
It has disappeared. Because the drink dispenser labeled `Mr. Right' is visible on the left counter in the first and second retrieved images but not in the third retrieved image or the current image. The inconsistency in the third sub-answer may be due to a different shooting time that failed to capture the dispenser. This suggests that the drink dispenser was indeed there before but has since disappeared.

\end{tcolorbox}

\caption{A case of inconsistent intermediate reasoning.}
\label{tab:casestudy1_inconsistent}
\end{table*}

\end{document}